\documentclass{article}

\PassOptionsToPackage{numbers, compress}{natbib}
\usepackage[final]{neurips_2019}

\usepackage{microtype}
\usepackage{graphicx}
\usepackage{subfigure}
\usepackage{booktabs} % for professional tables
\usepackage{mathtools}

\usepackage{ownstyles}
\usepackage{amsmath}
\usepackage{tikz}

\usepackage{hyperref}
\usepackage[utf8]{inputenc} % allow utf-8 input
\usepackage[T1]{fontenc}    % use 8-bit T1 fonts
\usepackage{hyperref}       % hyperlinks
\usepackage{url}            % simple URL typesetting
\usepackage{booktabs}       % professional-quality tables
\usepackage{amsfonts}       % blackboard math symbols
\usepackage{nicefrac}       % compact symbols for 1/2, etc.
\usepackage{microtype}      % microtypography

\graphicspath{{figures/}}

\newcommand{\capa}{\textbf{(a)}\xspace}
\newcommand{\capb}{\textbf{(b)}\xspace}
\newcommand{\capc}{\textbf{(c)}\xspace}
\newcommand{\capd}{\textbf{(d)}\xspace}

\newcommand{\capleft}{\textbf{(left)}\xspace}

\newcommand{\capright}{\textbf{(right)}\xspace}

\newcommand{\capsec}[1]{\textbf{#1}\xspace}
\newcommand{\citenop}[1]{\citeauthor{#1} \citep{#1}\xspace}   

\newcommand{\defeq}{\vcentcolon=}

\usepackage{array}
\newcolumntype{L}{>{\centering\arraybackslash}m{2.5cm}}

\newif\ifcomments

\commentsfalse
\ifcomments
\newcommand{\comments}[1]{#1}
\else
\newcommand{\comments}[1]{}
\fi

\newcommand{\titl}{LCA: Loss Change Allocation for \\Neural Network Training}

\title{\titl}

\author{%
  Janice Lan\\
  Uber AI \\
  \texttt{janlan@uber.com} \\
  \And
  Rosanne Liu \\
  Uber AI \\
  \texttt{rosanne@uber.com} \\
  \And
  Hattie Zhou \\
  Uber \\
  \texttt{hattie@uber.com} \\
  \And
  Jason Yosinski \\
  Uber AI \\
  \texttt{yosinski@uber.com} \\
}

\begin{document}

\maketitle

\begin{abstract}
  Neural networks enjoy widespread use, but many aspects of their training, representation, and operation are poorly understood.
  In particular, our view into the training process is limited, 
  with a single scalar loss being the most common viewport into this high-dimensional, dynamic process.
  %with the most commonly used measurement providing visibility into this high-dimensional, dynamic process being a scalar loss value.
  % We propose a new method for partitioning loss 
  % Despite the fact that neural networks have become very widely used, the training process generally remains a black box beyond observing the loss and confirming that it decreases.
  We propose a new window into training called Loss Change Allocation (LCA), in which credit for changes to the network loss is conservatively partitioned to the parameters.
  This measurement is accomplished by decomposing the components of an approximate path integral along the training trajectory using a Runge-Kutta integrator.
  % method using Runge-Kutta over the per-parameter gradient of the loss of the entire data set to gain insights into the training process.
  This rich view shows which parameters are responsible for decreasing or increasing the loss during training, or which parameters ``help'' or ``hurt'' the network's learning, respectively.
  LCA may be summed over training iterations and/or over neurons, channels, or layers for increasingly coarse views.
  %The views afforded by this new measurement device produce several insights into training.
  This new measurement device produces several insights into training.
  (1) We find that barely over 50\% of parameters help during any given iteration.
  (2) Some entire layers hurt overall, moving on average against the training gradient, a phenomenon we hypothesize may be due to phase lag in an oscillatory training process.
  (3) Finally, increments in learning proceed in a synchronized manner across layers, often peaking on identical iterations.
  % on a microscopic level, learning peaks in a synchronized manner across layers.
\end{abstract}

%%%%%%%%%%%%%%%%%%%%%%%%%%%%%%%%%%%%%%%%%%%%%%%%%%%%%%%%%%%%%%%%%%%%%
%%%%%%%%%%%%%%%%%%%%%%%%%%%%%%%%%%%%%%%%%%%%%%%%%%%%%%%%%%%%%%%%%%%%%
\vspace*{-0.5em}
\section{Introduction}
\seclabel{introduction}
\vspace*{-0.5em}

In the common stochastic gradient descent (SGD) training setup, a parameterized model is iteratively updated using gradients computed from mini-batches of data chosen from some training set.
Unfortunately, our view into the high-dimensional, dynamic training process is often limited to watching a scalar loss quantity decrease over time.
%decreases to the loss---or when things don't go as well, per-parameter responsibility for increases to the loss.
%
%\todo{maybe: add more motivation} \rl{Done. See below} We know how optimization methods work theoretically, and we have some insights about the where the parameters end up, but we don't really know much about how the parameters get there.
%
%\todo{(Maybe brief mentions of some related work here?)} \rl{Done. See below} Prior work focuses on explaining the final state of neural networks after training. And those that study the training dynamics typically only look at layers, not individual parameters.
%
%\rl{my attempt to rewrite the above:  $\scriptstyle\wedge\scriptstyle\wedge\scriptstyle\wedge$}
%\jby{nice!}
%
%We begin by briefly reviewing related work.
There has been much research attempting to understand neural network training, with some work studying
geometric properties of the objective function \cite{goodfellow2014qualitatively, li2018visualizing, soudry2016no, safran2016quality, nguyen2017loss},
properties of whole networks and individual layers at convergence
%We have gained certain understanding about where parameters end up when converge (e.g. local minima that are nearly globally optimal
\cite{choromanska2015loss, goodfellow2014qualitatively, keskar2016large, zhang2019all},
and 
neural network training from an optimization perspective \cite{sutskever2013importance, choromanska2015loss, dauphin2014identifying, bottou2018optimization, li-2018-ICLR-measuring-the-intrinsic-dimension}.
This body of work in aggregate provides rich insight into the loss landscape arising from typical combinations of neural network architectures and datasets.
%post-training behaviors of networks \cite{keskar2016large, goodfellow2014qualitatively} and layers \cite{zhang2019all}.
Literature on the dynamics of the training process itself is more sparse, but a few salient works examine
the learning phase through the diagonal of the Hessian, mutual information between input and output, and other measures
%\cite{sutskever2013importance}: also studies learning dynamics (``transient'' and ``minimization'' phases)}
\cite{achille-2017-arXiv-critical-learning-periods,shwartz-ziv-2017-arXiv-opening-the-black-box-of-deep,jastrzebski-2019-ICLR-on-the-relation-between-the-sharpest}.

%\cite{achille-2017-arXiv-critical-learning-periods} dynamics in two phases
%\cite{shwartz-ziv-2017-arXiv-opening-the-black-box-of-deep} two phases: ``has two different and distinct phases: empirical error minimization (ERM) and representation compression''

%\rl{cites: 
%``The Loss Surfaces of Multilayer Networks" -- local minima is like global minima.
%``Identifying and attacking the saddle point problem in high-dimensional non-convex optimization'' -- saddle points exist but it's ok.
%``Qualitatively characterizing neural network optimization problems'' -- local minima don't matter.
%``Online learning and stochastic approximations'' -- SGD converges to local minima where expected loss is 0.
%``Global Optimality in Neural Network Training'' -- local minima are globally optimal and a local descent strategy can reach a global minima from any initialization.
%``On the Quality of the Initial Basin in Overspecified Neural Networks'' -- under some conditions 
%High probability of initializing at a point from which there is a monotonically decreasing path to a global minimum;
%}

In this paper we propose a simple approach to inspecting training in progress by decomposing changes in the overall network loss into a per-parameter \emph{Loss Change Allocation} or \emph{LCA}.
The procedure for computing LCA is straightforward, but to our knowledge it has not previously been employed for investigating network training.
We begin by defining this measure in more detail, and then apply it to reveal several interesting properties of neural network training. Our contributions are as follows:

%With the ability to calculate per-parameter, per-iteration Loss Change Allocation, the rest of this paper focuses on employing it to visualize (\secref{visualizing}) and understand network training properties, leading to three conclusions learned about networks:

%Our contributions in this paper are:
\begin{enumerate}
% \vspace*{-0.2em}
\item We define the Loss Change Allocation as a per-parameter, per-iteration decomposition of changes to the overall network loss (\secref{approach}). Exploring network training with this measurement tool uncovers the following insights.
  
\item Learning is very noisy, with only slightly over half of parameters helping to reduce loss on any given iteration (\secref{noise}).
%  Using this tool, we can see how surprisingly noisy training is, and we garner some insights into why that's the case

\item Some \emph{entire layers} consistently drift in the wrong direction during training, on average moving \emph{against} the gradient. 
%We give a plausible explanation in terms of these layers being slightly out of phase during training (\secref{hurtinglayers}). 
We propose and test an explanation that these layers are slightly out of phase, lagging behind other layers during training (\secref{hurtinglayers}).

\item We contribute new evidence to 
%\removed{the discussion concerning the ``top-down'' vs ``bottom-up'' views of training. Evidence suggests}
suggest that the learning progress is, on a microscopic level, \emph{synchronized} across layers, with small peaks of learning often occurring at the same iteration for all layers (\secref{synchronized}).
%  We show that layers learn concepts in a synchronized manner, (contrary to "bottom-up" or "top-down" frameworks in prior works).
%\vspace*{-1em}
\end{enumerate}

%%%%%%%%%%%%%%%%%%%%%%%%%%%%%%%%%%%%%%%%%%%%%%%%%%%%%%%%%%%%%%%%%%%%%
%%%%%%%%%%%%%%%%%%%%%%%%%%%%%%%%%%%%%%%%%%%%%%%%%%%%%%%%%%%%%%%%%%%%%

\vspace*{-1em}
\section{The Loss Change Allocation approach}
\seclabel{approach}
\vspace*{-0.5em}

\figp[t]{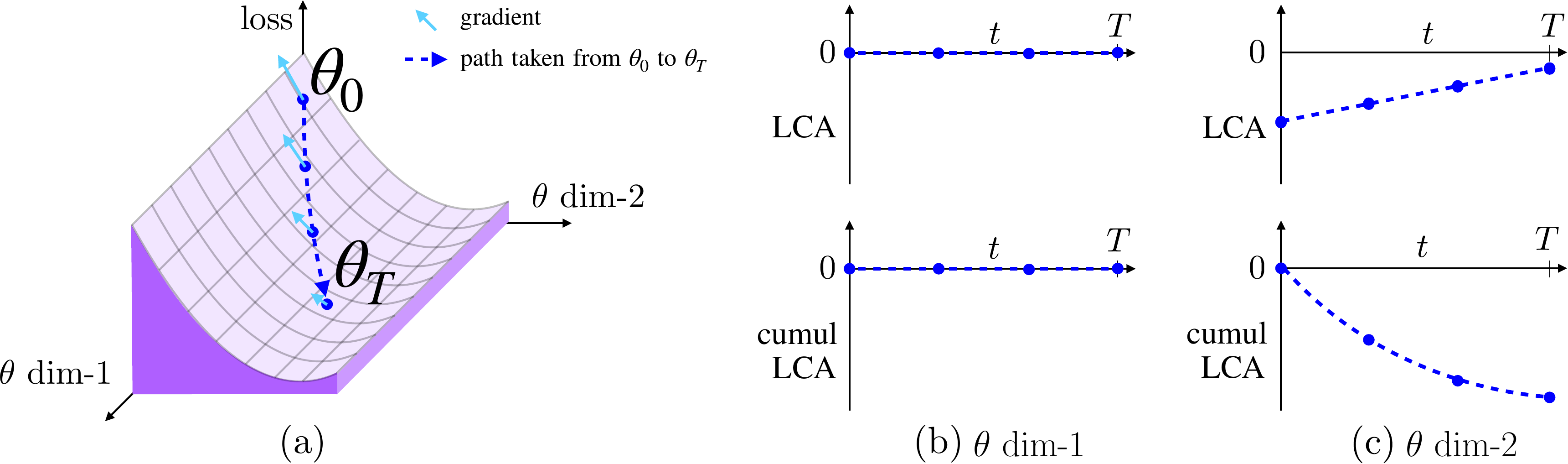}{.82}{
  \capa
  Illustration of this paper's method on a toy two-dimensional loss surface.
  %defined as a function of a two-dimensional parameter vector $\theta$.
  We allocate credit for changes to the model's training loss to individual parameters \capb $\theta~\mathrm{dim\mbox{-}1}$ and \capc $\theta~\mathrm{dim\mbox{-}2}$
  by multiplying parameter motion with the corresponding individual component of the gradient of the training set.
  %The gradient of the whole training set is expensive to compute but on modern hardware not prohibitively so.
  This partitions changes to the loss into individual \emph{Loss Change Allocation (LCA)} components
  allows us to measure which parameters learn at each timestep, providing a rich view into the training process. In the example depicted, although both parameters move, the second parameter captures all the credit, as only its component of the gradient is non-zero.
  %\todo{maybe combine with H matrix diagram?}
  \vspace*{-2em}
}

\newcommand{\dtrain}{\ensuremath{\mathcal{D}_{\mathrm{train}}}\xspace}
\newcommand{\batcht}{\ensuremath{\mathcal{B}_t}\xspace}

We begin by defining the Loss Change Allocation approach in more detail.
Consider a parameterized training scenario where a model starts at parameter value $\theta_0$ and ends at parameter value $\theta_T$ after training. The training process entails traversing some path $P$ along the surface of a loss landscape from $\theta_0$ to $\theta_T$. There are several loss landscapes one might consider; in this paper we analyze the \emph{training} process, so we measure motion along the loss with respect to the \emph{entire training set}, here denoted simply $L(\theta)$. We analyze the loss landscape of the training set instead of the validation set because we aim to measure training, not training confounded with issues of memorization vs. generalization (though the latter certainly should be the topic of future studies).

The approach in this paper derives from a straightforward application of the fundamental theorem of calculus to a path integral along the loss landscape:

\vspace*{-1em}
\beq
L(\theta_T) - L(\theta_0) =
\int_C \langle \nabla_\theta L(\theta), d\theta \rangle
\eeq
\vspace*{-1em}

where $C$ is any path from $\theta_0$ to $\theta_T$ and $\langle\cdot{,}\cdot\rangle$ is the dot product. This equation states that the change in loss from $\theta_0$ to $\theta_T$ may be calculated by integrating the dot product of the loss gradient and parameter motion along a path from $\theta_0$ to $\theta_T$. Because $\nabla_\theta L(\theta)$ is the gradient of a function and thus is a conservative field, any path from $\theta_0$ to $\theta_T$ may be used; in this paper we consider the path taken by the optimizer during the course of training.
We may approximate this path integral from $\theta_0$ to $\theta_T$ by using a series of first order Taylor approximations along the training path. If we index training steps by $t \in [0, 1, ..., T]$, the first order approximation for the change in loss during one step of training is the following, rewritten as a sum of its individual components:

\vspace*{-1em}
\begin{align}
  L(\theta_{t+1}) - L(\theta_{t})
  & \approx \langle \nabla_\theta L(\theta_{t}),~\theta_{t+1}-\theta_{t} \rangle \eqnlabel{one} \\
  & = \sum_{i=0}^{K-1} (\nabla_\theta L(\theta_{t}))^{(i)}(\theta_{t+1}^{(i)}-\theta_{t}^{(i)})
  %\defeq \sum_{i=0}^{K-1} C^{(i)}
  \defeq \sum_{i=0}^{K-1} A_{t,i}
\eqnlabel{two}
\end{align}
\vspace*{-1em}

where $\nabla_\theta L(\theta_{t})$ represents the gradient of the loss of the whole training set w.r.t. $\theta$ evaluated at $\theta_{t}$,
$v^{(i)}$ represents the $i$-th element of a vector $v$, and
the parameter vector $\theta$ contains $K$ elements.
Note that while we \emph{evaluate model learning} by tracking progress along the training set loss landscape $L(\theta)$, training itself is accomplished using stochastic gradient approaches in which noisy gradients from mini-batches of data drive parameter updates via some optimizer like SGD or Adam.
%Note that the update to $\theta$ is determined using the gradient of loss of the mini-batch, whereas $\nabla_\theta L(\theta_{t})$ is evaluated using the entire training set.
As shown in \eqnref{two}, the difference in loss produced by one training iteration $t$ may be decomposed into $K$ individual \emph{Loss Change Allocation}, or \emph{LCA}, components, denoted $A_{t,i}$. These $K$ components represent the LCA for a single iteration of training, and over the course of $T$ iterations of training we will collect a large $T \times K$ matrix of $A_{t,i}$ values.

%or a higher order integrator at each step
%with initial training set loss $L(\theta_a)$ and over the course of training moves to $\theta_b$ with hopefully lower loss $L(\theta_b)$.
%The change in loss as the 

%Consider a training setup where a model's parameter vector is updated each iteration using gradients from a mini-batch of data.
%The gradient of this mini-batch loss at $\theta_t$, along with any other adjustments such as momentum \joel{a bit imprecise}, generates an updated parameter vector $\theta_{t+1}$. We want to see how this movement (resulting from the mini-batch update) affects the loss aggregated across  the whole training set. The total loss presumably decreases in general, but which parameters had a greater impact on loss? One way to measure each parameter's effect on loss at step $t$ is to use a multivariate Taylor approximation:

The total loss over the course of training will often decrease, and the above decomposition allows us to allocate credit for loss decreases on a per-parameter, per-timestep level.
Intuitively, when the optimizer increases the value of a parameter and its component of the gradient on the whole training set is negative, the parameter has a negative LCA and is ``helping'' or ``learning''. Positive LCA is ``hurting'' the learning process, which may result from several causes: a noisy mini-batch with the gradient of that step going the wrong way, momentum, or a step size that is too large for a curvy or rugged loss landscape as seen in \cite{jastrzebski-2019-ICLR-on-the-relation-between-the-sharpest, xing2018walk}.
If the parameter has a non-zero gradient but does not move, it does not affect the loss. Similarly, if a parameter moves but has zero gradient, it does not affect the loss.
The sum of the $K$ components is the overall change in loss at that iteration.
\figref{drawing_per_param} depicts a toy example using two parameters. Throughout the paper we use ``helping'' to indicate negative LCA (a contribution to the reduction of total loss), and ``hurting'' for positive LCA.

An important property of this decomposition is that it is \emph{grounded}:
the sum of individual components equals the total change in loss, and each contribution has the same fundamental units as the loss overall (e.g. nats or bits in the case of cross-entropy). This is in contrast to approaches that measure quantities like parameter motion or approximate elements of the Fisher information (FI) \cite{kirkpatrick-2017-PNAS-overcoming-catastrophic-forgetting,achille-2017-arXiv-critical-learning-periods}, which also produce per-parameter measurements but depend heavily on the parameterization chosen. For example, the FI metric is sensitive to scale (e.g. multiply one relu layer weights by 2 and next by 0.5: loss stays the same but FI of each layer changes and total FI changes). Further, LCA has the benefit of being \emph{signed}, allowing us to make measurements and interpretations when training goes backwards (Sections \ref{sec:noise} and \ref{sec:hurtinglayers}). 
% LCA is beneficial vs FIM because it is grounded to the loss and signed. Grounded: While [1] and [13] are illuminating, the FIM metric used may be rescaled arbitrarily (e.g. multiply one relu layer weights by 2 and next by 0.5: loss stays the same but FIM of each layer changes and total FIM changes). In contrast LCA is grounded, so its units are the same as loss (e.g. bits or nats for cross entropy) and its scale is fixed. Signed: FIM is unsigned and thus could not yield our “50.7\% help” nor “some layers move backward” conclusion.

%With \eqnref{two}, we can decompose the change in loss into $K$ components, one component per parameter. We use this measure as the basis of our results and analyses.

Ideally, summing up the $K$ components should equal $L(\theta_{t+1}) - L(\theta_{t})$. In practice, the first order Taylor approximation is often inaccurate due to the curvature of the loss landscape. 
We can improve on our LCA approximation from \eqnref{one} by replacing
$\nabla_\theta L(\theta_{t})$
with
$\frac{1}{6} ( \nabla_\theta L(\theta_{t}) + 4 \nabla_\theta L(\frac{1}{2}\theta_{t} + \frac{1}{2}\theta_{t+1}) + \nabla_\theta L(\theta_{t+1}) )$, with the $(1, 4, 1)$ coefficients coming from the fourth-order Runge–Kutta method (RK4) \cite{runge-1895-uber-die-numerische-auflosung,kutta-1901-beitrag-zur-naherungweisen-integration} or equivalently from Simpson's rule \cite{weisstein-2003-simpsons-rule}. Using a midpoint gradient doubles computation but shrinks accumulated error drastically, from first order to fourth order.
If the error is still too large, we can halve the step size with composite Simpson's rule by calculating gradients at $\frac{3}{4}\theta_{t} + \frac{1}{4}\theta_{t+1}$ and $\frac{1}{4}\theta_{t} + \frac{3}{4}\theta_{t+1}$ as well. We halve the step size until the absolute error of change in loss per iteration is less than 0.001, and we ensure that the cumulative error at the end of training is less than 1\%.
First order and RK4 errors can be found in \tabref{fo_rk_errors} in Supplementary Information.
Note that the approach described may be applied to
any parameterized model trained via gradient descent, but for the remainder of the paper we assume the case of neural network training.

%%%%%%%%%%%%% Fig moved for layout vvvvvvvvv
\figgsp[t]{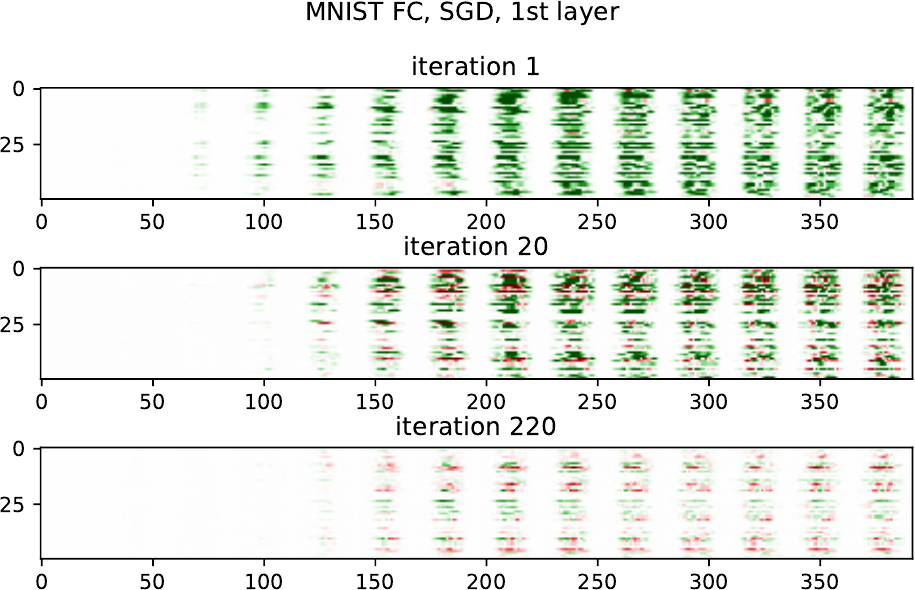}{0.48}{3frames_lenet.pdf}{0.48}{
    Frames from an animation of the learning process for two training runs. \capsec{(left)} The 1st layer of an MNIST--FC (full  shape is 100$\times$784, but only the upper left quarter is shown for better clarity). \capsec{(right)} The 2nd convolutional layer of an MNIST--LeNet (full shape is 40$\times$20 of 5$\times$5 blocks; only upper left quarter is shown). Each pixel represents one parameter. The LeNet layer shows 5$\times$5 grids representing each filter, laid out by input channels (columns) and output channels (rows).
    Parameters that help (decrease the loss) at a given time are shown as shades of green. Parameters that hurt (increase the loss) are shown as shades of red. Larger magnitudes of LCA are darker and white indicates zero LCA.
    % JBY: removed vvv this because the visualization is fine without it.
    % Within each network, the same color and shade represents the same LCA value across the different iterations.
  Iteration 20 is partly through the main drop in loss, and 220 is one full epoch.
  In MNIST--FC, we can see clusters spaced at intervals of 28 pixels, because these parameters connect to the flattened MNIST images.
  %White bands between clusters correspond to parameters that do not learn because their input pixels are never non-zero (a peculiarity of MNIST).
  Learning is strongest in early iterations with mostly negative LCA, remains strong for many iterations but with more variance in LCA across parameters, and has greatly diminished by iteration 220, where much of learning is complete.
  The complete animations may be viewed at: \url{https://youtu.be/xcnoRnoVyXQ} and \url{https://youtu.be/EY3LoXmdkYU}.
  \vspace*{-1.7em}
}
%%%%%%%%%%%%% Fig moved for layout ^^^^^^^^^^

\subsection{Experiments}
\seclabel{experiments}
%\vspace*{-.5em}

We employ the LCA approach to examine training on two tasks: MNIST and CIFAR-10, with architectures including a 3-layer fully connected (FC) network and LeNet \cite{lecun-1998-IEEE-gradient-based-learning-applied} on MNIST, and AllCNN \cite{springenberg-2014-arXiv-striving-for-simplicity:-the-all-convolutional} and ResNet-20 \cite{he-2015-arXiv-deep-residual-learning} on CIFAR-10. Throughout this paper we refer to training runs as ``dataset--network'', e.g., \textbf{MNIST--FC}, \textbf{MNIST--LeNet}, \textbf{CIFAR--AllCNN}, \textbf{CIFAR--ResNet}, followed by further configuration details (such as the optimizer) when needed.

%
%For MNIST, we use a 3-layer fully connected (FC) network. %No batch norm or dropout was used. We also used 
%as well as LeNet \cite{lecun-1998-IEEE-gradient-based-learning-applied}. %for an example of convolution, with two conv2D layers with max pooling followed by two fully connected layers with dropout \cite{hinton2012improving-neural-networks-by-preventing}.
%For CIFAR, we use AllCNN \cite{springenberg-2014-arXiv-striving-for-simplicity:-the-all-convolutional} with 9 convolutional layers followed by global average pooling. Batch normalization~\cite{ioffe-2015-arXiv-batch-normalization:-accelerating} and dropout~\cite{hinton2012improving-neural-networks-by-preventing} are used. We also use a deep residual network, ResNet-20 \cite{he-2015-arXiv-deep-residual-learning}.
%%as described in [\url{https://github.com/keras-team/keras/blob/master/examples/cifar10_resnet.py} and cite \url{https://arxiv.org/pdf/1512.03385.pdf}].

For each dataset--network configuration, we train with both SGD and Adam optimizers, and conduct multiple runs with identical hyperparameter settings. 
Momentum of 0.9 is used for all SGD runs, except for one set of ``no-momentum'' MNIST--FC experiments. Learning rates are manually chosen between 0.001 to 0.5. See \secref{si:experiment} in Supplementary Information for more details on architectures and hyperparameters. We also make our code available at \url{https://github.com/uber-research/loss-change-allocation}.
Note that we use standard network architectures to demonstrate use cases of our tool; we strive for simplicity and interpretability of results rather than state-of-the-art performance. Thus we do not incorporate techniques such as L2 regularization, data augmentation, and learning rate decay.
Since our method requires calculating gradients of the loss over the entire training set, it is considerably slower than the regular training process, but remains tractable for small to medium models; see \secref{comptime} for more details on computation.

\subsection{Direct visualization}
\seclabel{visualizing}

We calculate LCA for every parameter at every iteration and animate the LCA values through all the iterations in the whole training process. \figref{3frames_fc.pdf} shows snapshots of frames from the video visualization. In such videos, we arrange parameters first by layer and then for each layer as two-dimensional matrices (1-D vectors for biases), and overlay LCA values as a heatmap.
This animation enables a granular view of the training process.

We can also directly visualize each parameter versus time, granting each parameter its own training curve. We can optionally aggregate over neurons, channels, layers, etc. (see \secref{si:visualization} for examples).
A benefit of these visualizations is that they convey a large volume of data directly to the viewer, surfacing subtle patterns and bugs that can be further investigated.
Observed patterns also suggest more quantitative metrics that surface traits of training. The rest of the paper is dedicated to such metrics and traits.

%%%%%%%%%%%%% Fig moved for layout vvvvvvvvvv
\figp[t]{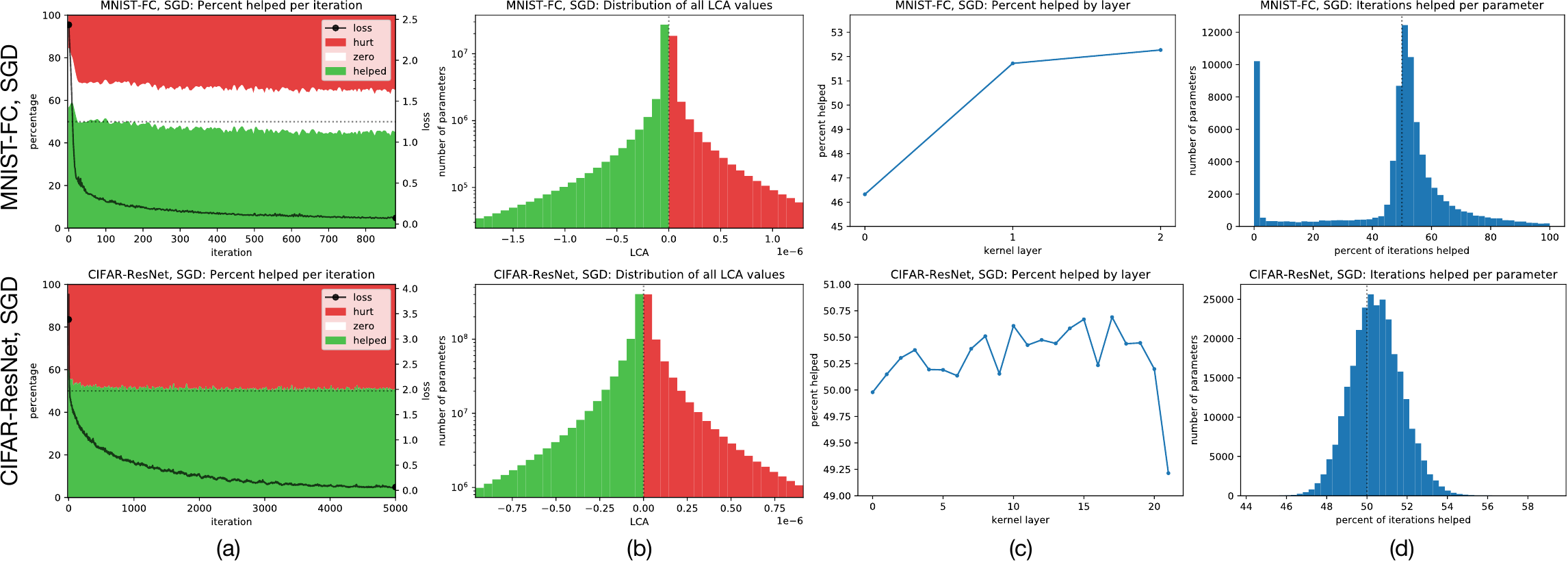}{.9}{
    \capa Visualization of the percentage of parameters that helped, hurt, or had zero effect through training, overlaid with the loss curve of that run.
    \capb The distribution of helping and hurting LCA (zeros ignored) over the entire training, zoomed in to ignore 1\% of tails. 
    \capc Average percent of weights helping for each layer in network, curiously near 50\% for all.
    \capd Histogram of the fraction of iterations each weight helped, showing that most weights swing back and forth between helping and hurting evenly.
 In every column the first row is MNIST--FC and second row CIFAR--ResNet, both trained with SGD. 
  Notable facts:
  MNIST--FC shows a significant percent of weights with zero effect. Because MNIST has pixels that are never on, any first layer weights connected to those pixels cannot help or hurt. 
  %Counts by LCA bin are balanced.
  %
  CIFAR--ResNet exhibits barely over 50\% of parameters helping over the course of training, even during the period of significantly learning (loss reduction) from iteration 0 to 2000. Averaged over the entire run, only 50.66\% of parameters helped (see \tabref{percenthelped}).
  % PUT BACK IN IF TRUE:
  % \secref{si:noise} shows this plot for other networks, showing similar behavior.
  %
  Note that in both runs we can see that in the earliest iterations, the percent of weights helping is higher, but only slightly.
%We can see in this two example runs that earlier iterations have a slightly higher percentage of weights helping (60\% for MNIST--FC and 62\% for CIFAR--ResNet). After that they quickly decrease to an average around 50\%.
% this was done using the first run of may16_allnetworks/mnistfc and i believe first run of jan30/resnetsgd...
 \vspace*{-1em}
}
%%%%%%%%%%%%% Fig moved for layout ^^^^^^^^^^

%%%%%%%%%%%%%%%%%%%%%%%%%%%%%%%%%%%%%%%%%%%%%%%%%%%%%%%%%%%%%%%%%%%%%%%%%%%%%%%%%%%%%%%%%
% Results 1: barely more than half help
%%%%%%%%%%%%%%%%%%%%%%%%%%%%%%%%%%%%%%%%%%%%%%%%%%%%%%%%%%%%%%%%%%%%%%%%%%%%%%%%%%%%%%%%%
\vspace*{-0.5em}
\section{Learning is very noisy}
\seclabel{noise}
\vspace*{-0.5em}
%One direct observation from visualizing per-parameter, per-iteration LCA data (\figref{3frames_fc.pdf}) is that many parameters hurt instead of help on a given iteration. Rather than all parameters slowly pitching in their micro-scale Loss Change Allocations to lower the overall loss, there is significant fighting between parameters.
%It may be understandable as parameters are updated according to minibatch gradient which is itself a noisy estimation of the overall gradient.

%However, it might still be surprising to find out that on average, \emph{almost half of parameters are hurting}, in every training iteration. Moreover, each parameter, including ones that help in total, \emph{hurt almost almost half of the time}. 

Although it is a commonly held view that the inherent noise in SGD-based neural network training exists and is even considered beneficial \cite{keskar2016large}, this noise is often loosely defined as a deviation in gradient estimation. While the minibatch gradient serves as as a suggested direction for parameter movement, it is still one step away from the actual impact on decreasing loss over the whole training set, which LCA represents precisely. 
%In contrast, LCA allows the understanding of noise through the more straightforward lens of loss movement, from a population point of view. \joel{The previous two sentences could benefit from a bit more clarity -- why is it a more straightforward lens, what do mean by a deviation in gradient estimation.}
By aggregating a population of per-parameter, per-iteration LCAs along different axes, we 
present numerical results that shed light into the noisy learning behavior.
%\removed{Note that the measured properties about training dynamics change with network configuration including the problem (e.g. MNIST or CIFAR), architecture (e.g. FC or CNN), and optimizer (e.g. SGD or Adam), and other hyperparameters, and we generally try to separate these factors, and conduct multiple independent runs under each configuration.}
We find it surprising that on average \emph{almost half of parameters are hurting} in every training iteration. Moreover, each parameter, including ones that help in total, \emph{hurt almost half of the time}.
%\jl{I decided not to bother with "what might our mental model be" here since we started talking about noise already, but maybe we could include that...}
%\jby{ok}

\vspace*{-.5em}
\paragraph{Barely over 50\% of parameters help during training.}

\begin{table}[b]
  \caption{Percentage of helping parameters (ignoring those with zero LCA) for various networks and optimizers, averaged across all iterations and 3 independent runs per per configuration.
  }
\tablabel{percenthelped}
% \vskip 0.15in
\begin{center} \begin{small}
    %\begin{sc}
%\setlength\tabcolsep{1.5pt}
\begin{tabular}{lccccr}
\toprule
 & MNIST-FC, mom=0 & MNIST-FC & MNIST-LeNet & CIFAR-ResNet & CIFAR-AllCNN \\
\midrule
SGD  & $53.72 \pm 0.05$ & $57.79 \pm 0.16$  & $53.97 \pm 0.48$  & $50.66 \pm 0.14$   & $51.09 \pm 0.23$ \\
Adam & N/A              & $55.82 \pm 0.09$  & $51.77 \pm 0.21$  & $50.30 \pm 0.004$  & $50.19 \pm 0.01$ \\
\bottomrule
\end{tabular}
%  \end{sc}
\end{small} \end{center}
\vskip -0.1in
\end{table}

According to our definition, for each iteration of training, parameters can help, hurt, or not impact the overall loss. With that in mind, we count the number of parameters that help, hurt, or neither, across all training iterations and for various networks; two examples of networks are shown in \figref{phpi_combined_crop} (all other networks shown in \secref{si:noise}).
The data show that in a typical training iteration, close to half of parameters are helping and close to half are hurting! This ratio is slightly skewed towards helping in early iterations but stays fairly constant during training. 
Averaged across all iterations,
the percentage of helping parameters for various network configurations is reported in \tabref{percenthelped}. 
We see that it varies within a small range of 50\% to 58\%, with CIFAR networks even tighter at 50\% to 51\%. This observation also holds true when we look at each layer separately in a network; \figref{phpi_combined_crop}(c) shows that all layers have similar ratios of helpful parameters.

\vspace*{-.5em}
\paragraph{Parameters alternate helping.} Now that we can tell if each parameter is ``helpful'', ``hurtful'', or ``neither''\footnote{We rarely see ``neither'', or zero-impact parameters in CIFAR networks, but it can be of a noticable amount for MNIST (around 20\% for MNIST--FC; see \figref{phpi_combined_crop}), mostly due to the many dead pixels in MNIST.}, we wonder if parameters predictably stay in the same category throughout training. In other words, is there a consistent elite group of parameters that always help?
When we measure the percentage of helpful iterations per parameter throughout a training run, histograms in \figref{phpi_combined_crop}(d) show that parameters help approximately half of the time, and therefore the training of a network is achieved by parameters alternating to make helpful contribution to the loss. 

Additionally, we can measure the oscillations of individual parameters. 
\figref{oscillate_weights_resnet_crop} shows a high number of oscillations in weight movement for CIFAR--ResNet on SGD: on average, weight movements change direction once every 6.7 iterations, and gradients change signs every 9.5 iterations. \secref{si:noise} includes these measures for all networks, as well as detailed views in \figref{valleys_combined_crop} suggesting that many of these oscillations happen around local minima.
While oscillations have been previously observed for the overall network \cite{xing2018walk, jastrzebski-2019-ICLR-on-the-relation-between-the-sharpest},
thanks to LCA, we're able to more precisely quantify the individual and net effects of these oscillations. As we'll see in \secref{hurtinglayers}, we can also use LCA to identify when a network is damaged not by oscillations themselves, but by their precise phase.
% JBY: I took the below out in favor of the reworded version above.
%1. oscillations happen on a parameter level, 2. they are ubiquitous across layers and for various optimizers with or without momentum, 3. they cause parameters to alternate between helping and hurting.

%\rl{Probably makes sense to combine \figref{phpi_combined_crop} and \figref{placeholder_perc_helped_layer_crop.pdf}, and mark each figure group with (a) (b) (c) (d). left and right are not ideal when they are referred across different paragraphs}
%\done{combined figs}

% OLD FIG; Moved to be combined with phpi_combined_crop.pdf!
%\figgp{placeholder_perc_helped_layer_crop.pdf}{0.4}{helped_hist_resnet.pdf}{0.45}{
%  \capleft Average percent of weights helped within a layer. Shown is a CIFAR--ResNet trained with SGD. Shaded areas are stdevs over 10 runs?? \later{make percents out of 100}
%  \capright Histogram of percentage of helping iterations for every parameter. Every parameter helps approximately half the time.
%  \todo{What is this plot? And what should we see here?}
%  \secref{si:noise} shows this plot for other networks, showing similar behavior.
%  \todo{yeah let's combine with fig 3}
%}

\vspace*{-.5em}
\paragraph{Noise persists across various hyperparameters.}
Changing the learning rate, momentum, or batch size (within reasonable ranges such that the network still trains well) only have a slight effect on the percent of parameters helping. See \secref{si:noise} for a set of experiments on CIFAR--ResNet with SGD, where percent helped always stays within 50.3\% to 51.6\% for reasonable hyperparameters.

\vspace*{-.5em}
\paragraph{Learning is heavy-tailed.}
A reasonable mental model of the distribution of LCA might be a narrow Gaussian around the mean. However, we find that this is far from reality. Instead, the LCA of both helping and hurting parameters follow a heavy-tailed distribution, as seen in \figref{phpi_combined_crop}(b).
\figref{heavy_tail_redo_crop.pdf} goes into more depth in this direction, showing that contributions from the tail are about three times larger than would be expected if learning were Gaussian distributed.
More precisely, a better model of LCA would be the Weibull distribution with $k<1$. 
The measurements suggest that the view of learning as a Wiener process \cite{shwartz-ziv-2017-arXiv-opening-the-black-box-of-deep} should be refined to reflect the heavy tails.

%In both helping and hurting groups, the amount of LCA of all parameters and iterations follow a heavy-tailed distribution. Histograms of the same two example networks can be seen in . It shows that most parameters impact the loss very little, and learning is driven by a few strong players (not the same ones throughout training). \figref{heavy_tail_redo_crop.pdf} shows the distribution within small time intervals throughout training, to help visualize the change across time, and compare it with a simulated Gaussian distribution.
%
%Given this observation, a random walk or Brownian motion would not be the right way to think about neural network training; it is closer to a Levy flight or a Wiener process.

%%%%%%%%%%%%%%%%%%%%%%%%%%%%%%%%%%%%%%%%%%%%%%%%%%%%%%%%%%%%%%%%%%%%%%%%%%%%%%%%%%%%%%%%%
% Results 2: some layers hurt
%%%%%%%%%%%%%%%%%%%%%%%%%%%%%%%%%%%%%%%%%%%%%%%%%%%%%%%%%%%%%%%%%%%%%%%%%%%%%%%%%%%%%%%%%

\vspace*{-0.5em}
\section{Some layers hurt overall}
\seclabel{hurtinglayers}
\vspace*{-0.5em}

Although our method is used to study low-level, per-parameter LCA, we can also aggregate these over higher level breakdowns for different insights; individually there is a lot of noise, but on the whole, the network learns. 
The behavior of individual layers during learning has been of interest to many researchers \cite{zhang2019all,raghu-2017-arXiv-svcca:-singular-vector}, so a simple and useful aggregation is to sum LCA over all parameters within each layer and sum over all time, measuring how much each layer contributes to total learning.

We see an expected pattern for MNIST--FC and MNIST--Lenet (all layers helping; \figref{layer_totals_combined_crop}), but CIFAR--ResNet with SGD shows a surprising pattern: the \emph{first and last layers consistently hurt training} (positive total LCA). Over ten runs, the first and last layer in ResNet hurt statistically significantly
%p-values of 1e-6 and 3e-5 respectively,
(p-values $< 10^{-4}$ for both),
whereas all other layers consistently help (p-values $< 10^{-4}$ for all). Blue bars in \figref{resnet_lastlayer_crop.pdf} shows this distinct effect. Such a surprising observation calls for further investigation.
%We thus designed a number of experiments to moderate layer's learning specifically. 
The following experiments shed light on why this might be happening.

\figp[t]{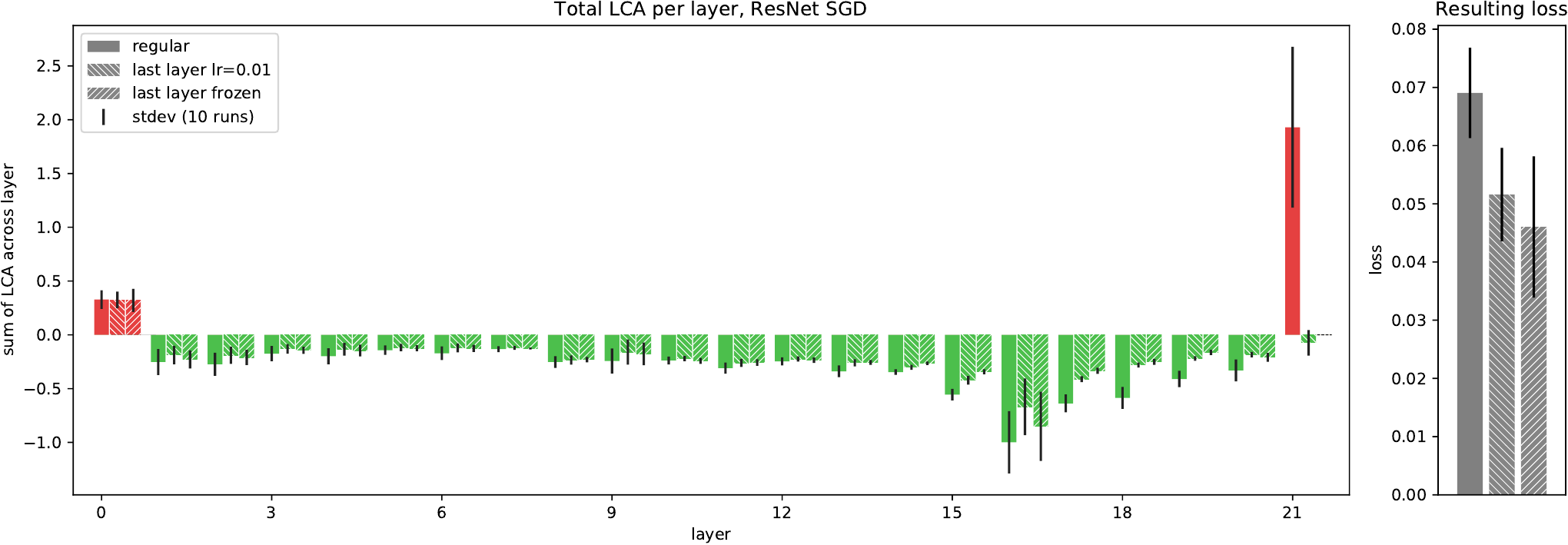}{0.95}{
  \capleft LCA summed over all of training, for each layer, in CIFAR--ResNet trained with SGD. Bias and batch norm layers are combined into their corresponding kernel layers. Blue represents regular runs. Orange is with the last layer frozen at initialization. Note that the other layers, especially the adjacent few, do not help as much, but the difference in LCA of the last layer is greater than the total differences of the other layers helping less. Green is with the last layer at a 10x smaller learning rate than the rest of the network, showing similar layer LCAs as when the layer is frozen.
  \capright Resulting train loss and standard deviations for each run configuration.
  Means and standard deviations are over 10 runs for each experiment configuration. %For each of those 10 runs, the same weight initialization and random mini-batch shuffle order were used in the various experiments.
\vspace*{-1em}
}

% old, with more details about the first layer
% First, we try freezing the first layer at its initialization. Though the first layer is no longer able to hurt, overall performance is not any better (insignificantly worse) because the other layers, especially the neighboring ones, don't help as much.
% We observed that the cumulative trajectory of the layer's LCA dips before becoming increasingly positive (\figref{resnet_freeze_combined_crop.pdf}(b)), so we also try letting the layer learn normally in the beginning and then freezing it at its LCA argmin. However, that also doesn't improve performance. Perhaps the shape of the layer trajectory represents the two phases found in \cite{shwartz-ziv-2017-arXiv-opening-the-black-box-of-deep}; the forgetting phase is just as important as the first phase even if it might hurt certain layers. From these results, the first layer of a ResNet appears to be an important sacrificial layer.
\vspace*{-0.5em}
\paragraph{Freezing the first layer stops it from hurting but causes others to help less.}
We try various experiments freezing the first layer at its random initialization. Though we can prevent this layer from hurting, the overall performance is not any better because the other layers, especially the neighboring ones, start to help less; see \figref{resnet_firstlayer_crop.pdf} for details. Nonetheless, this can be useful for reducing compute resources during training as you can freeze the first layer without impairing performance. 
%\done{ Joel: Meta note at this point: Maybe it is explicitly stated, but if we plan on using the somewhat colloquaial ``layer X hurt'' as opposed to ``layer X hurt loss'', we should state that explicitly somewhere.}

\vspace*{-0.5em}
\paragraph{Freezing the last layer results in significant improvement.}
In contrast to the first layer, freezing the last layer at its initialization (\figref{resnet_lastlayer_crop.pdf}) improves training performance (and test performance curiously; not shown), with p-values $< 0.001$ for both train loss and test loss, over 10 runs! We also observe other layers, especially neighboring ones, not helping as much, but this time the change in the last layer’s LCA more than compensates. Decreasing the learning rate of the last layer by 10x (0.01 as opposed to 0.1 for other layers) results in similar behavior as freezing it.
%\joel{Interesting result! Maybe could be accentuated more, as it is pretty counter-intuitive -- why would freezing the last layer improve test performance?}
These experiments are consistent with findings in \cite{hoffer2018lastlayer} and \cite{gotmare2018heuristics}, which demonstrate that you can freeze the last layer in some networks without degrading performance. With LCA, we are now able to provide an explanation for when and why this phenomenon happens. The instability of the last layer at the start of training in \cite{gotmare2018heuristics} can also be measured by LCA, as the LCA of the last layer is typically high in the first few iterations.
%\cite{hoffer2018lastlayer} and \cite{gotmare2018heuristics} have previously discovered that you can freeze the last layer in some networks without degrading performance. With LCA, we are now able to provide an explanation for when and why this happens. 

% We use a 10x smaller learning rate for just the first layer. It looks almost exactly the same as regular runs. We do the same for the last layer. It's almost like freezing the last layer. Huh.
% \todo{trend of last layer starting out helping and then hurting more and more}
\vspace*{-0.5em}
\paragraph{Phase shift hypothesis: is the last layer phase-lagged?}
While it is interesting to see that decreasing the learning rate by 10x or to zero changes the last layer's behavior, this on its own does not explain why the layer would end up going \emph{backwards}. The mini-batch gradient is an unbiased estimator of the whole training set gradient, so on average the dot product of the mini-batch gradient with the training set gradient is positive. Thus we must look beyond noise and learning rate for explanation.
We hypothesize that the last layer may be \emph{phase lagged} with respect to other layers during learning. Intuitively, it may be that while all layers are oscillating during learning, the last layer is always a bit behind. As each parameter swings back and forth across its valley, the shape of its valley is affected by the motion of all other parameters. If one parameter is frozen and all other parameters trained infinitesimally slowly, that parameters valley will tend to flatten out. This means if it had climbed a valley (hurting the loss), it will not be able to fully recover the LCA in the negative direction, as the steep region has been flattened. If the last layer reacts slower than others, its own valley walls may tend to be flattened before it can react.

\figgp[t]{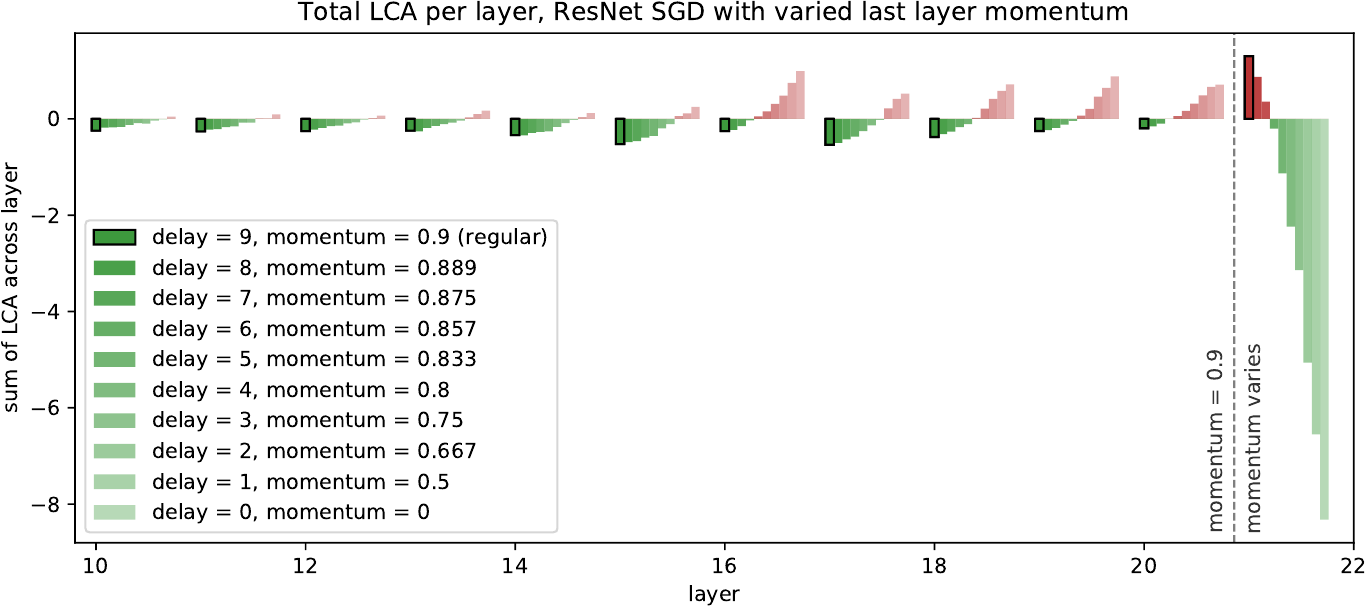}{0.64}{controllers_lc_vs_mom_crop.pdf}{0.35}{
  CIFAR--ResNet SGD with varying momentum for the last layer (and a fixed 0.9 for all other layers). Selected momentum values are derived from linear values of delay $[0, 1, 2, ..., 9]$ in a control system, where $\mathrm{momentum} = \mathrm{delay} / (\mathrm{delay} + 1)$, and a delay of 9 corresponds to regular runs of 0.9 momentum.
  \capleft LCA per layer (only the second half of the network is shown for better visibility; first half follows a similar trend, but less pronounced). As the last layer helps more, the other layers hurt more because they are relatively more delayed. \capright LCA of the last layer is fairly linear with respect to the delay.
\vspace*{-1em}
}

A simple test for this hypothesis is as follows. We note that training with momentum 0.9 corresponds to an information lag of 9 steps (the mean of an exponential series with exponent .9)---each update applied uses information 9 steps old. To give the last layer an advantage, we train it with momentum corresponding to a delay of $n$ for $n \in \{9, 8, ..., 0\}$ while training all other layers as usual. As shown in \figref{controllers_all_delays_redgreen_crop.pdf}, this works, and the transition from hurting to helping (a lot) is almost linear with respect to delay!
As we give the last layer an information freshness advantage, it begins to ``steal progress'' from other layers, eventually forcing the neighboring layers into positive LCA.
%Previously we've observed that weights oscillate, swinging between positive and negative LCA, and that the first and last layers have slightly more frequent oscillations (\figref{oscillate_weights_resnet_crop}). However, oscillations alone do not explain why the total would be positive or negative. We hypothesize that there is some sort of phase shift between the weight movement and the gradients. Based on concepts in control theory \todo{Jason to fill in}, we experimented with various measures of information delay for the last layer by adjusting momentum. \figref{controllers_all_delays_viridis_crop.pdf} supports our hypothesis. When the last layer is delayed less than the rest of the network, it is able to help more. With the right phase shifts, we can influence how much different layers help or hurt.
%
These results suggest that it may be profitable to view training as a fundamentally oscillatory process upon which much research in phase-space representations and control system design may come to bear.

Beyond CIFAR--Resnet, other networks also show intriguingly heterogeneous layer behaviors. 
As we noted before, in the case of MNIST--FC and MNIST--LeNet trained with SGD, all layers help with varying quantities.
An MNIST--ResNet (added specifically to see if the effect we see above is due to the data or the network) shows the last layer hurting as well. We also observe the last layer hurting for CIFAR--AllCNN with SGD (\figref{allcnn_lastlayer_crop}) and multiple layers hurting for a couple of VGG-like networks (\figref{rebuttal_mini_vgg_lca_crop}).
When using Adam instead of SGD, CIFAR--ResNet has a consistently hurting first layer and an inconsistently hurting last two layers. CIFAR--AllCNN trained with Adam does not have any hurting layers.
We note that layers hurting is not a universal phenomenon that will be observed in all networks, but when it does occur, LCA can identify it and suggest potential candidates to freeze. Further, viewing training through the lens of information delay seems valid, which suggests that per-layer optimization adjustments may be beneficial.

\vspace*{-0.5em}
\section{Learning is synchronized across layers}
\seclabel{synchronized}
\vspace*{-0.5em}

We learned that layers tend to have their own distinct, consistent behaviors regarding hurting or helping from per-layer LCA summed across all iterations. In this section we further examine the per-layer LCA \emph{during} training, equivalent to studying individual ``loss curves'' for each layer, and discover that the exact moments where learning peaks are curiously synchronized across layers. And such synchronization is not driven by only gradients or parameter motion, but both.

%\todo{need new transition} The reader may have noticed in the prior section that when a particular class was learned (say, the 2 that was learned last), the learning was accomplished by contributions from different layers at around the same iteration. The relative ratios changed over time (bottom early, top late), but the differently sized spikes occurred at about the same training iteration.

%Spurred by this observation, in this section we measure the extent to which learning is synchronized across layers. 
We define ``moments of learning'' as temporal spikes in an instantaneous LCA curve, local minima where loss decreased more on that iteration than on the iteration before or after, and show the top 20 such moments (highest magnitude of LCA) for each layer in \figref{fc_aggregate_worm_plot_vf_highlight_crop}. We further decompose this metric by class (10 for both MNIST and CIFAR), where the same moments of learning are identified on per-class, per-layer LCAs, shown in \figref{fc_worm_plot_vf_highlight_crop}. Whenever learning is synchronized across layers (dots that are vertically aligned) they are marked in red. Additional figures on CIFAR--ResNet can be seen in \secref{si:synchronized}.
The large proportion of red aligned stacks suggests that learning is very locally synchronized across layers.

To gauge statistical significance, we compare the number of synchronized moments in these networks to a simple baseline: the number we would observe if each layer had been randomly shifted to one or two iterations earlier or later.
We find that the number synchronized moments is significantly more than that of such a baseline (p-value $< 1^{-6}$). See details on this experiment in \secref{si:synchronized}.
Thus, we conclude that for these networks we've measured, learning happens curiously synchronously across layers throughout the network.
We might find different behavior in other architectures such as transformer models or recurrent neural nets, which could be of interest for future work.
%though (per the previous section) with ratios that may change over time.

But what drives such synchronization? Since learning is defined as the product of parameter motion and parameter gradient, we further examine whether one of them is synchronized in the first place. By plotting in the same fashion of identified local peaks, we observe the synchronization pattern in gradients per layer is clearly different from that in LCA, either in terms of the total loss (\figref{fc_aggregate_worm_plot_eff_grads_vf_highlight_crop}) or per-class loss (\figref{fc_worm_plot_eff_grads_vf_highlight_crop}). Since parameter motion (\figref{fc_aggregate_worm_plot_weights_vf_highlight_crop}) is the same across all classes, it alone doesn't drive the per-class LCA. We therefore conclude that
the synchronization of learning, demonstrated by synchronized behavior in LCA movement (\figref{fc_worm_plot_vf_highlight_crop}), is strong, and comes from both parameter motion and gradient.

\figp[t]{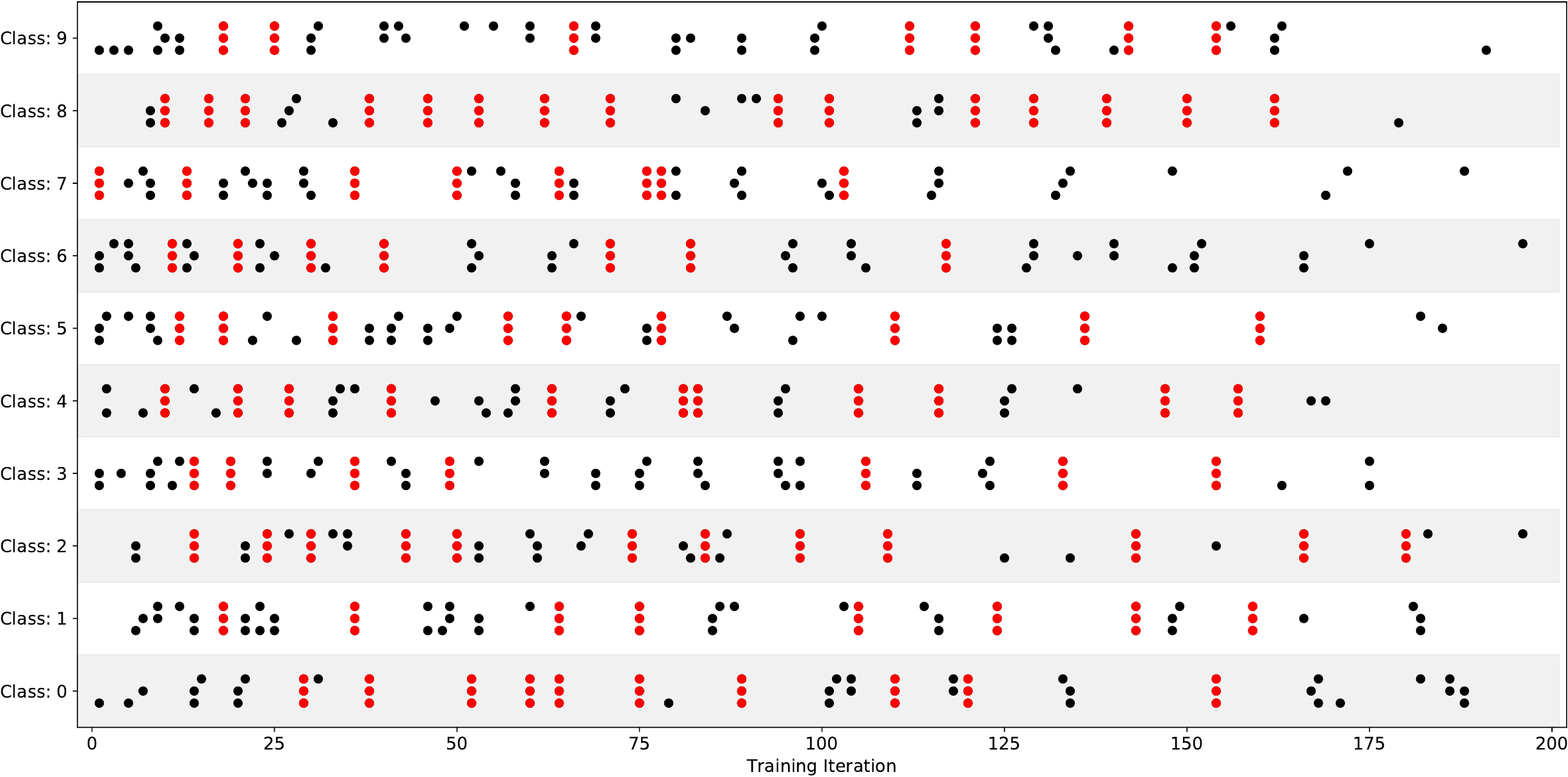}{.9}{
    Peak learning iterations by layer by class on MNIST--FC. The same LCA data as in \figref{fc_aggregate_worm_plot_vf_highlight_crop} but seperated by class.
  We plot the top 20 iterations by LCA for each class and each layer, where that iteration represents a local minimum for LCA. The layers are ordered from bottom to top. Points highlighted in red represent iterations where all three layers had peak learning for that particular class. To measure the statistical significance of these vertical line structures in red, we simulate a baseline by shifting each layer in each class randomly by -2, -1, 0, or 1, 2 iteration. We find that the average number of vertical lines is 0.4 in the baseline and 9.4 in the actual network, and this difference is significant with a p-value < 0.001.
%\vspace*{-1em}
}

%%%%%%%%%%%%%%%%%%%%%%%%%%%%%%%%%%%%%%%%%%%%%%%%%%%%%%%%%%%%%%%%%%%%%%%%%%%%%%%%%%%%%%%%%
% (ENTIRE SECTION MOVED TO SI) Results 4: A 3 layer story
%%%%%%%%%%%%%%%%%%%%%%%%%%%%%%%%%%%%%%%%%%%%%%%%%%%%%%%%%%%%%%%%%%%%%%%%%%%%%%%%%%%%%%%%%

%%%%%%%%%%%%%%%%%%%%%%%%%%%%%%%%%%%%%%%%%%%%%%%%%%%%%%%%%%%%%%%%%%%%%
% Results last: misc
%%%%%%%%%%%%%%%%%%%%%%%%%%%%%%%%%%%%%%%%%%%%%%%%%%%%%%%%%%%%%%%%%%%%%
% \section{Other observations (probably put in SI?)}
% Correlation within outputs

% Cannot predict the future

%%%%%%%%%%%%%%%%%%%%%%%%%%%%%%%%%%%%%%%%%%%%%%%%%%%%%%%%%%%%%%%%%%%%%
%%%%%%%%%%%%%%%%%%%%%%%%%%%%%%%%%%%%%%%%%%%%%%%%%%%%%%%%%%%%%%%%%%%%%

\vspace*{1.5em}
\section{Conclusion}
%\vspace*{-0.5em}

The Loss Change Allocation method acts as a microscope into the training process, allowing us to examine the inner workings of training with much more fine-grained analysis. When applied to various tasks, networks and training runs, we observe many interesting patterns in neural network training that induce better understanding of training dynamics, and bring about practical model improvements.
%\todo{summarize key insights?}

\subsection{Related work}
\seclabel{related_work}
%\vspace*{-.5em}

We note additional connections to existing literature here.
%
%\paragraph{Per-parameter investigation}
%Per-parameter investigation into neural networks, in previous works, in addition to being scarce, concerns only the magnitude of weights.
The common understanding is that learning in networks is sparse; a subnetwork \cite{frankle-2019-ICLR-the-lottery-ticket-hypothesis}, or a random subspace of parameters \cite{li-2018-ICLR-measuring-the-intrinsic-dimension} 
is sufficient for optimization and generalization.
%The result of this line of work often supports network pruning. 
Our method provides an additional, more accurate, measure of usefulness to characterize per-parameter contribution.
A similar work \cite{zenke-2017-arXiv-improved-multitask-learning} defines per-parameter importance in the same vein but is computed locally with the mini-batch gradient, which overestimates the true per-parameter contribution to the decrease of loss of the whole training set.
%structural regularizer to alleviate catastrophic forgetting

%However, the basis of this type of investigation has been limited to the magnitude of weight values, lacking a better, or additional, view on characterizing parameter contributions.

%\paragraph{Understanding the training process}
Several previous works have increased our understanding of the training process.
\citenop{alain-2016-arXiv-understanding-intermediate-layers} measured and tracked over time the ability to linearly predict the final class output given intermediate layers representations.
\citenop{raghu-2017-arXiv-svcca:-singular-vector} found that networks converge to final \emph{representations} from the bottom up, and class-specific information in networks is formed at various places.
\citenop{shwartz-ziv-2017-arXiv-opening-the-black-box-of-deep} visualized the training process through the information plane, where two phases are identified as empirical error minimization of each followed by a slow representation compression. There measurements are developed but none have examine the process each individual parameter undergoes.

%\paragraph{Interesting findings}
Methods like saliency maps \cite{simonyan2013deep-inside-convolutional}, DeepVis
\cite{yosinski-2015-arXiv-understanding-neural-networks},
and others allow interpretation of representations or loss surfaces.
%\todo{cite some things} \cite{}.
But these works only approach the end result of the model, not the training process in progress. LCA can be seen as a new tool that specializes on the microscopic level of details, and such inspection follows through the whole training process to reveal interesting facts about learning. 
Some of our findings resonate with and complement other work. For example, in \cite{zhang2019all} it is also observed that layers have heterogeneous characteristics; in that work layers are denoted as either ``robust'' or ``critical'', and robust layers can even be reset to their initial value with no negative consequence.

%Potential newer works to cite:
%\begin{itemize}
%   \item "Are all layers created equal"
%   \item "critical learning periods in deep networks" (ICLR)
%   \item "on the relation between the sharpest directions of dnn loss and the sgd step length" (ICLR)
%\end{itemize}

\subsection{Future work}

There are many potential directions to expand this work.
%We could run the LCA method on more types of models and tasks.
Due to the expensive computation and the amount of analyses, we have only tested vision classification tasks on relatively small datasets so far. In the future we would like to run this on larger datasets and tasks beyond supervised learning, since the LCA method directly works on any parameterized model.
An avenue to get past the expensive computation is to analyze how well this method can be approximated with gradients of loss of a subset of the training set. We are interested to see if the observations we made hold beyond the vision task and the range of hyperparameters used. 

Since per-weight LCA can be seen as a measurement of weight importance, an simple extension is to perform weight pruning with it, as done in \cite{frankle-2019-ICLR-the-lottery-ticket-hypothesis,zhou2019deconstructing}
(where weight's final value is used as an importance measure). %Some initial experiments have already shown promising results.
Further, if there are strong correlations between underperforming hyperparameters and patterns of LCA, this may help in architecture search or identifying better hyperparameters.

%Our observation of each layer's distinct behavior suggest that layer-specific optimizers, hyperparameters could be beneficial.

%If there are strong correlations between underperforming hyperparameters and patterns of LCA, this may help in architecture search or identifying better hyperparameters. 
%We can conduct plenty of experiments on different hyperparameters. For instance, what happens when we drop the learning rate? Do larger networks allow more parameters to be sacrifices? Some initial experiments have shown that pruning parameters by LCA works with methods in \cite{frankle-2019-ICLR-the-lottery-ticket-hypothesis}, and there is much more to be explored in that area. Further, if there are strong correlations between underperforming hyperparameters and patterns of LCA, this may help in architecture search or identifying better hyperparameters. 

We are also already able to identify which layers or parameters overfit by comparing their LCA on the training set and LCA on the validation or test set, which motions towards future work on targeted regularization.
Finally, the observations about the noise, oscillations, and phase delays can potentially lead to improved optimization methods.

%%%%%%%%%%%%%%%%%%%%%%%%%%%%%%%%%%%%%%%%%%%%%%%%%%%%%%%%%%%%%%%%%%%%%
%%%%%%%%%%%%%%%%%%%%%%%%%%%%%%%%%%%%%%%%%%%%%%%%%%%%%%%%%%%%%%%%%%%%%

% Acknowledgements should only appear in the accepted version.
\section*{Acknowledgements}
We would like to acknowledge Joel Lehman, Richard Murray, and members of the Deep Collective research group at Uber AI for conversations, ideas, and feedback on experiments.

% Acks:
% Joel: conversations and feedback on experiments
% Richard: conversations and feedback

\clearpage

\bibliography{bibdesk,additional}
\bibliographystyle{plainnat}

%%%%%%%%%%%%%%%%%%%%%%%%%%%%%%%%%%%%%%%%%%%%%%%%%%%%%%%%%%%%%%%%%%%%%
%
% SUPPLEMENTARY INFORMATION
%
%%%%%%%%%%%%%%%%%%%%%%%%%%%%%%%%%%%%%%%%%%%%%%%%%%%%%%%%%%%%%%%%%%%%%

\clearpage

\renewcommand{\thesection}{S\arabic{section}}
\renewcommand{\thesubsection}{\thesection.\arabic{subsection}}

% Issue: Resetting counters breaks the refs to the sections/figures in SI.
\newcommand{\beginsupplementary}{%
        \setcounter{table}{0}
    \renewcommand{\thetable}{S\arabic{table}}%
        \setcounter{figure}{0}
    \renewcommand{\thefigure}{S\arabic{figure}}%
        \setcounter{section}{0}
}

\beginsupplementary

\onecolumn
\noindent\makebox[\linewidth]{\rule{\linewidth}{3.5pt}}

\begin{center}
    {\LARGE \bf Supplementary Information for:\\ \titl\par}
\end{center}
\noindent\makebox[\linewidth]{\rule{\linewidth}{1pt}}

\section{Supplementary results: Method}
\seclabel{si:method}

\begin{table}[h]
    \caption{Summary of errors of the LCA method with the Runge-Kutta method (RK4) used in our analyses, as well as the first order Taylor approximation (FO) for comparison.
    ``Total error'' is the percent error of total change in loss based on LCA vs. actual total change in loss over all iterations, where negative means that LCA gave a lower final loss.
    ``Average iteration error'' is the absolute error in one iteration, averaged over all iterations.
    Positive and negative errors at individual iterations could hypothetically cancel out somewhat when summed over all of training, though this is clearly not the case for first order, as it consistently and severely overestimates how much the loss decreases. 
    This conforms with observations of the loss landscape being biased toward positive curvature.
    Reported numbers are averaged over 3 runs per configuration.}
\tablabel{fo_rk_errors}
\vspace*{-1.5em}
\begin{center} \begin{small} 
% \begin{tabular}{  p{3cm}  p{2cm}  p{2cm} p{2cm} p{2cm}} 
\begin{tabular}{lccLL}
\toprule
 & Total error, RK4 & Total error, FO & Average iteration error, RK4 & Average iteration error, FO \\
\midrule
MNIST-FC, SGD mom=0 & -0.27\% & -4249.64\%   & 6.11E-05 & 0.11400    \\
MNIST-FC, SGD       & 1.08E-03\% & -171.10\% & 2.94E-06 & 0.00459    \\
MNIST-FC, Adam      & 3.64E-03\% & -75.31\%  & 2.31E-06 & 0.00199    \\
MNIST-LeNet, SGD    & 1.87E-02\% & -274.84\% & 3.45E-06 & 0.00838    \\
MNIST-LeNet, Adam   & 0.0262\% & -498.67\%   & 2.30E-06 & 0.01499    \\
CIFAR-ResNet, SGD   & 0.0345\% & -1189.71\%  & 1.37E-05 & 0.01018    \\
CIFAR-ResNet, Adam  & 0.0537\% & -923.88\%   & 1.07E-05 & 0.00793    \\
CIFAR-AllCNN, SGD   & 0.1374\% & -1371.69\%  & 4.51E-06 & 0.00662    \\
CIFAR-AllCNN, Adam  & 0.0745\% & -972.61\%   & 2.77E-06 & 0.00482    \\
\bottomrule
\end{tabular} \end{small} \end{center} \end{table}

\section{Supplementary results: Direct visualization}
\seclabel{si:visualization}

Examples of additional direct visualization methods shown in \figref{si_lenetadam_neurons_crop.pdf} and \figref{si_lenetadam_params_00_crop.pdf}. \secref{si:learn_together} shows other possible visualizations with aggregations over neurons or channels.

\figp{si_lenetadam_neurons_crop.pdf}{0.9}{
  First layer of MNIST--LeNet with Adam. Rather than displaying all parameters in one iteration, we can sum up parameters within each output channel of this layer. The other axis can now be used to display all iterations. 
}

\figp{si_lenetadam_params_00_crop.pdf}{0.9}{
  Cumulative LCA for individual parameters in the first layer of MNIST--LeNet with Adam. Left: the 50 (out of 500) most helpful parameters (most negative LCA). Right: a random set of 50 parameters. You can see that the typical parameter's cumulative LCA drops quickly near the beginning and then continues to wiggle slightly after flattening out. Some parameters are mostly flat after the initial drop, but others continue learning slightly until the end.
}

%%%%%%%%%%%%%%%%%%%%%%%%%%%%%%%%%%%%%%%%%%%%%%%%%%%%%%%%%%%%%%%%%%%%%%%%%%%%%%%

\section{Supplementary results: Learning is very noisy}
\seclabel{si:noise}

% noise in general

We provide plots from \secref{noise} for all networks here in \figref{combined_fc_crop.pdf}, \figref{combined_lenet_crop.pdf}, \figref{combined_resnet_crop.pdf}, and \figref{combined_allcnn_crop.pdf}.

% parameters alternate
Plots of oscillation shown in \figref{oscillate_weights_resnet_crop} and \figref{valleys_combined_crop} for ResNet, and additional oscillation measurements in \tabref{frequencies} for all networks.

% noise persists
Adjusting hyperparameters has some effect on the percent of parameters helping, shown in \figref{hyperp_lr_crop.pdf}. However, the percent helped remains within a small range, especially when ignoring points of significantly worse test performance (rightmost points for momentum and mini-batch size).

\figp[h!]{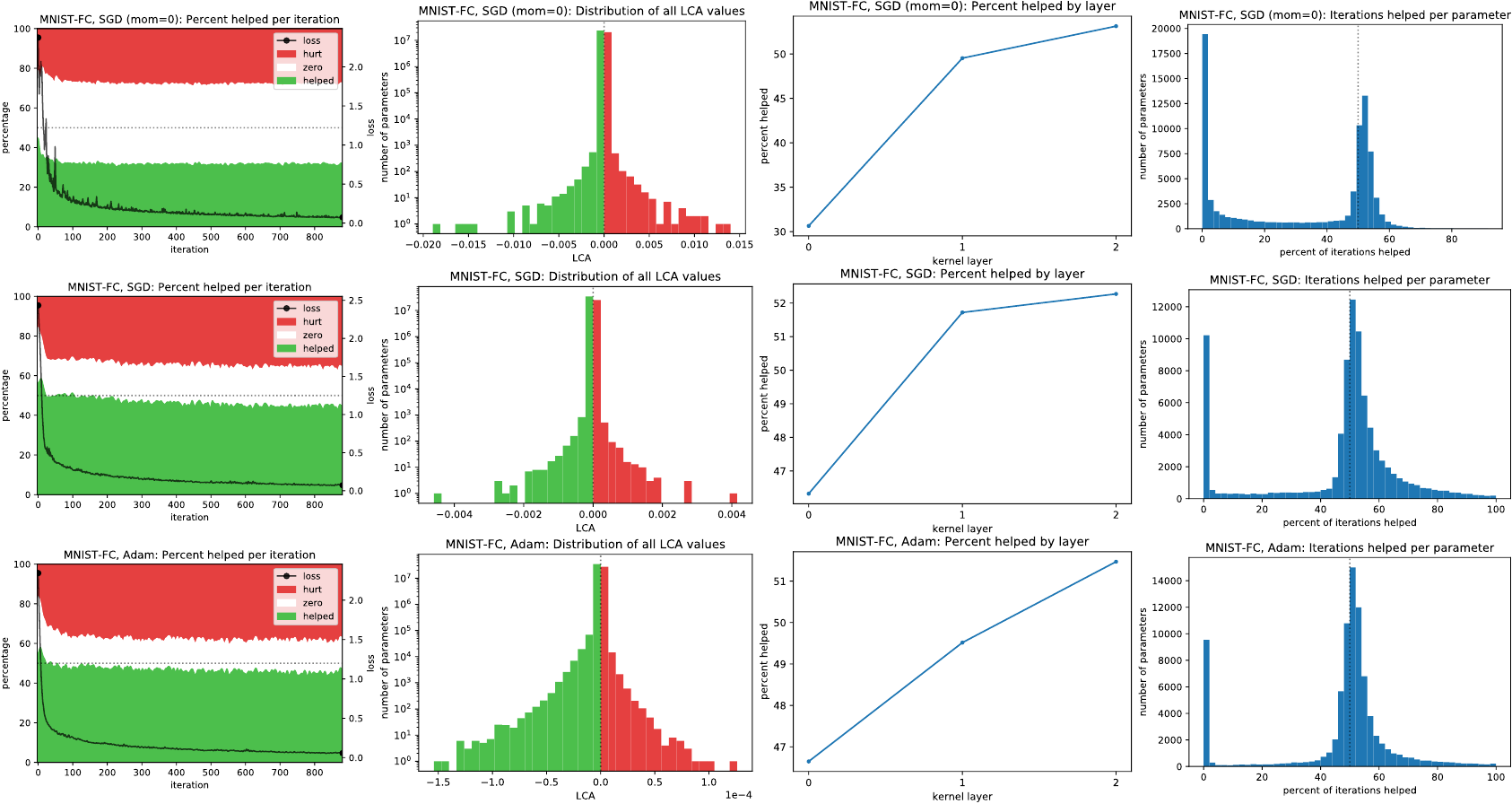}{1}{
  MNIST--FC. Top: SGD with no momentum, middle: SGD with momentum = 0.9, bottom: Adam. Figures display the same measurements as in \figref{phpi_combined_crop}, except the histogram of all LCA values now shows all values rather than ignoring the 1\% tails. 
}

\figp[h!]{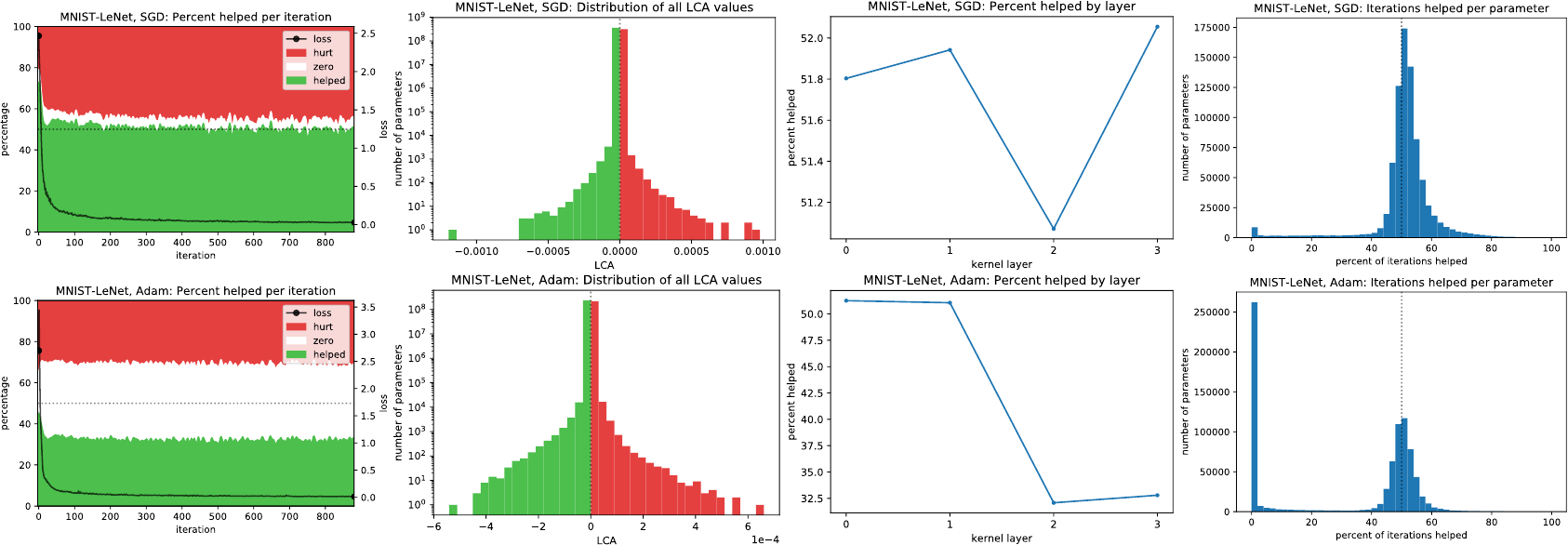}{1}{
  MNIST--LeNet. Top: SGD, bottom: Adam
}

\figp[h!]{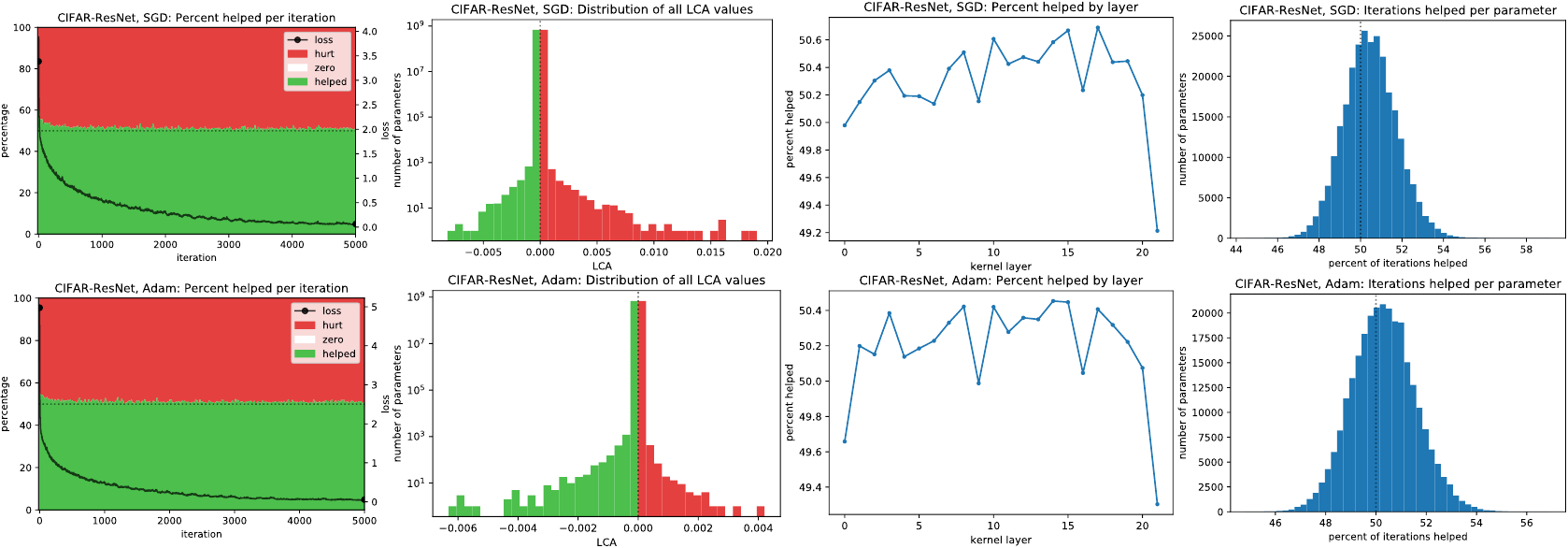}{1}{
  CIFAR--ResNet. Top: SGD, bottom: Adam
}

\figp[h!]{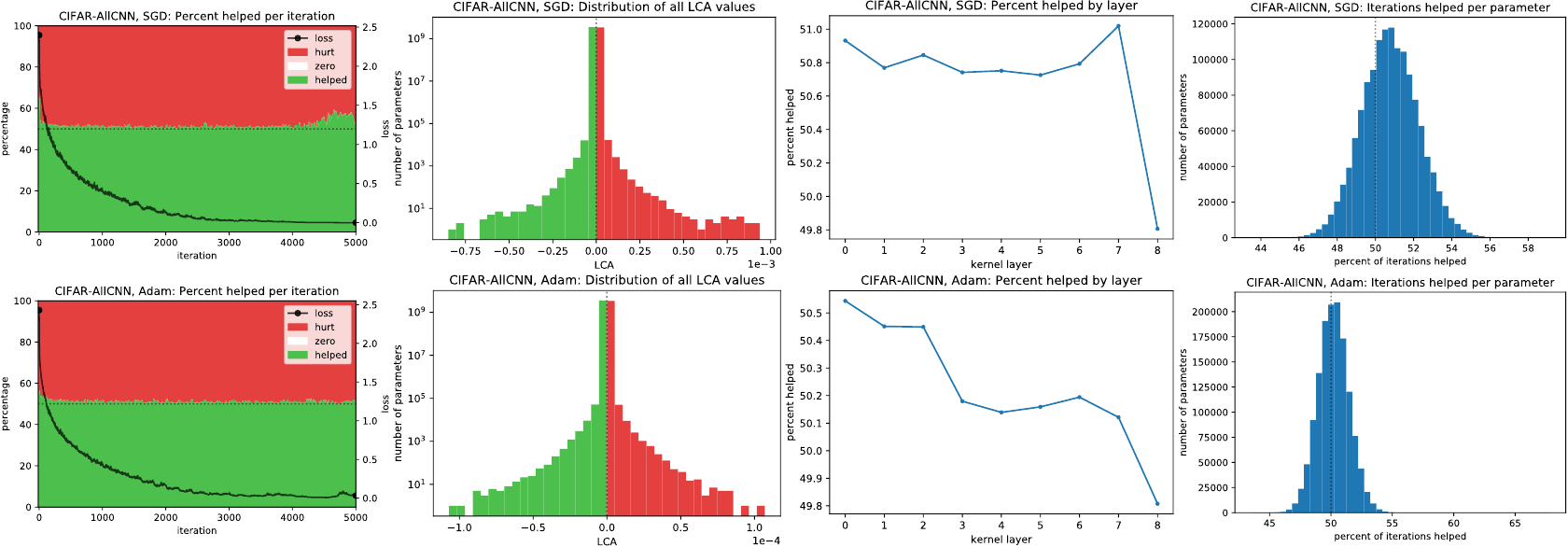}{1}{
  CIFAR--AllCNN. Top: SGD, bottom: Adam
}

\figgp{oscillate_weights_resnet_crop}{0.45}{oscillate_grads_resnet_crop}{0.45}{
  Oscillations for CIFAR--ResNet with SGD. Left: number of times the weight movement oscillates. Right: number of times the gradient crosses zero from one iteration to the next. Values shown are number of iterations out of 5000, averaged over all parameters within a layer. The averages over the entire network are 741.9 for weight turns and 525.8 for gradients crossing zero. Note that the first and last layers oscillate more than their neighboring layers, which is interesting given that those layers hurt (\secref{hurtinglayers}), but this is only a correlation as oscillations do not explain why something would bias towards helping or hurting.
}

% resnet gradient and weight same shape plots
% selected: may1_resnetfreeze/resnet_sgd0.1normal_1, last kernel layer, top 2 params that hurt the most in iterations 50 to 200. Ignored first 50 because  Param 149 and 548.

\figp{valleys_combined_crop}{1}{
  Oscillations of two individual parameters of the last kernel layer in CIFAR--ResNet with SGD. We look at iterations 50-200 (first 50 skipped due to large fluctuations), and display the parameter that hurt the most (top) and the parameter that helped the most (bottom) in these given iterations (net LCA of +3.41e-3 and -3.03e-3, respectively).
  Here, we visualize the weight movements (orange) along with their gradient values (blue) of the loss of the whole training set w.r.t. that parameter.
  Note that the trajectory of gradients and weights have similar shapes, indicating that these parameters are oscillating back and forth over a parabolic local minima that is shifting slightly (as the other parameters of the network are changing).
  During one back-and-forth cycle, the parameter will help, then hurt once the gradient crosses zero but momentum causes the weight to keep moving in the same direction, then help as the weight movement switches direction, and then hurt again when the gradient crosses zero again. This swinging LCA is depicted in the green and red bars.
  Because of these oscillations, the parameters end up helping approximately half the time and hurting the other half. While this behavior does not account for all the noise in other parameters and other iterations, it is commonly present.
}

\begin{table}
    \caption{Two metrics on oscillation: for each parameter, we look at the weight movement and count how often it switches direction (derivative of weight value changes sign) from one iteration to the next. We also look at the gradient for that parameter, and count how often it crosses zero (changes sign) from one iteration to the next. We convert both these into average frequencies over the training process, and then average those over all parameters in the network. These two measures are related -- if a parameter oscillates around a local minima, its gradient would cross zero every time the weight changes direction -- but due to noise, they do not have to correspond 1:1.
\rl{This is wrong?}}
\tablabel{frequencies}
\begin{center} \begin{small}
\begin{tabular}{  p{5cm}  p{3cm}  p{3cm} } 
\toprule
Network & Number of iterations per weight movement direction change & Number of iterations every time gradient crosses zero \\
\midrule
MNIST--FC, SGD mom=0  &  3.49  &  3.48  \\
MNIST--FC, SGD        &  13.57  &  11.68  \\
MNIST--FC, Adam       &  12.68  &  12.71  \\
MNIST--LeNet, SGD     &  10.29  &  9.37  \\
MNIST--LeNet, Adam    &  16.86  &  15.37  \\
CIFAR--ResNet, SGD    &  6.74  &  9.51  \\
CIFAR--ResNet, Adam   &  6.76  &  9.81  \\
CIFAR--AllCNN, SGD    &  7.06  &  11.18  \\
CIFAR--AllCNN, Adam   &  6.76  &  10.37  \\
\bottomrule
\end{tabular}
\end{small} \end{center}
\end{table}

\figggsp{hyperp_lr_crop.pdf}{0.3}{hyperp_mom_crop.pdf}{0.3}{hyperp_bs_updated_crop.pdf}{0.3}{
  Effects of different hyperparameter values on the percent of parameters helped per iteration, for CIFAR--ResNet with SGD. Left: learning rate, middle: momentum, right: mini-batch size. While there is some change in the percent helped, it does not go below 50.3\% or above 51.6\% within hyperparmeter ranges that still allow the network to learn. 
  This excludes the configuration of momentum = 0.99 with 53.5\% helped, where train and test performance have both degraded, and the number of parameters with zero LCA has actually significantly increased (resulting in 27.6\% helped, 48.5\% zero, 23.9\% hurt). Also, as we increase the mini-batch size to 5000 (1/10th of the training data), the percent helped reaches 56.2\%, but this is not practical as the test performance has become significantly worse.
}

% heavy tailed

% \figref{percentile_composite_crop} depicts how LCA distributions are heavy-tailed. In fact, it is significantly more heavy-tailed than a Gaussian: a kurtosis test on the \figref{percentile_composite_crop}(c) distribution against a Gaussian gives excess kurtosis of 10420 and a p-value of effectively 0. \figref{percentile_composite_crop}(d) has kurtosis of 2141, averaged across iteration intervals.

% old comparisons to gaussians
\figp[h!]{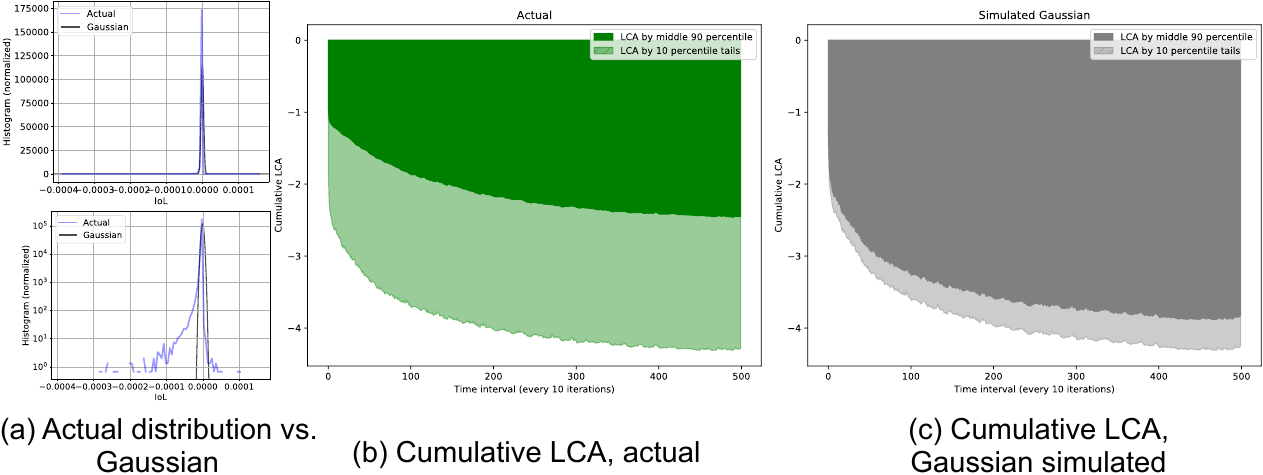}{0.9}{
  A depiction of how LCA distributions are significantly more heavy-tailed than a Gaussian distribution. A kurtosis test on the actual distribution against a Gaussian gives excess kurtosis of 10420 and a p-value of effectively 0. The actual distribution has kurtosis of 2141, averaged across iteration intervals
  \removed{Removed xmas bars}
  % LCA from inliers and outliers of the weight population. For every 100 iteration interval, we look at summed LCA made within that interval, by either the top and bottom 5\% (outliers), or the middle 90\% (inliers). Green bars indicate helped (negative values of LCA), and red hurt. \capa Solid bars are contributions made by inliers, and lighter color and hatched bars are those by outliers. \capb Results from the cancellation of positive and negative LCAs of the above, with respect to their class, to reveal how much actual LCA is from inliers versus outliers. \capc Actual distribution (of the first time interval) versus a simulated Gaussian with matching mean and variance. Bottom is in log space to manifest the tail difference.  
  % \capd A cumulative of (b), which truthfully recovers the loss curve of this training. \cape A comparison to simulated Gaussian distributions with the same mean and variance at every time interval, also divided into the same two classes.
}

%%%%%%%%%%%%%%%%%%%%%%%%%%%%%%%%%%%%%%%%%%%%%%%%%%%%%%%%%%%%%%%%%%%%%%%%%%%%%%%
\clearpage
\section{Supplementary results: Some layers hurt overall}
\seclabel{si:layerhurt}

Additional plots: \figref{layer_totals_combined_crop} comparing 3 different networks, \figref{rebuttal_mini_vgg_lca_crop} demonstrating the phenomenom of layers hurting for additional network architectures, \figref{resnet_firstlayer_crop.pdf} for experiments concerning the first layer of CIFAR--ResNet, SGD, and \figref{allcnn_lastlayer_crop} concerning the last layer of CIFAR--AllCNN, SGD.

\figp{layer_totals_combined_crop}{1}{
  LCA summed over all of training, across each layer. Bias and batch norm layers are combined into their corresponding kernel layers. Left: MNIST--FC, middle: MNIST--LeNet, right: CIFAR--ResNet, all using SGD. While there is variation in the FC and LeNet layers (magnitudes are somewhat correlated to the size of the layer), they all are helping with negative LCA. On the other hand, the first and last layers of the ResNet strangely have positive LCA. 
}

\figp[h!]{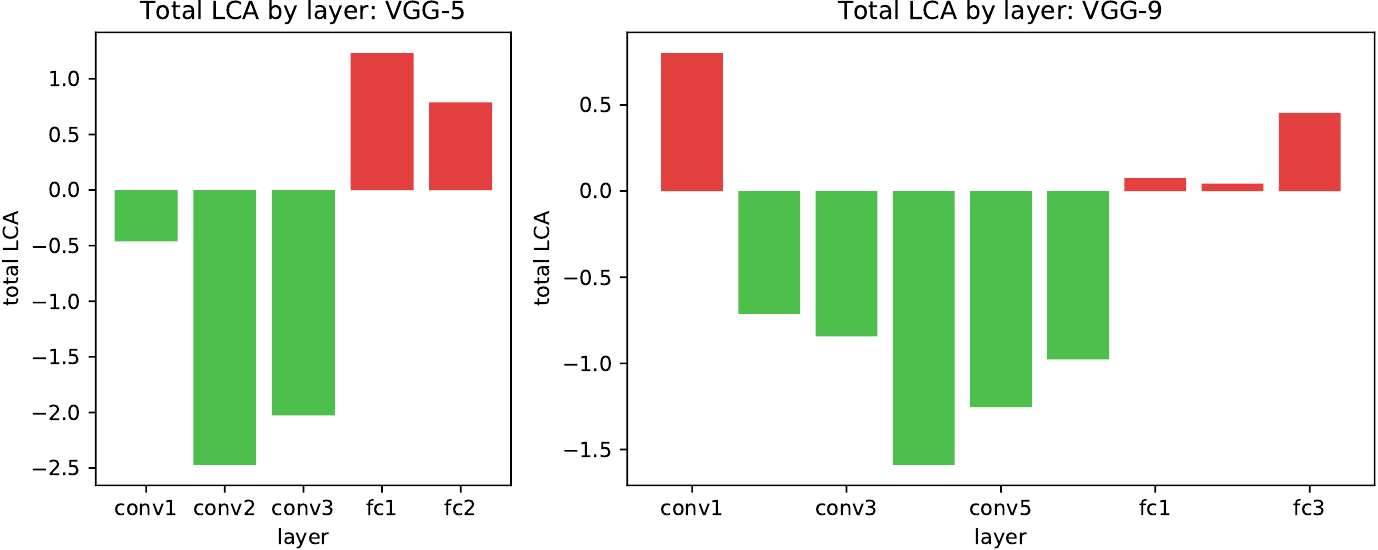}{0.9}{
  LCA of layers for two VGG-like models to further demonstrate that layers hurting is not a special one-off observation. It is also interesting to note that multiple layers hurt in these networks. Left: VGG-5 has 3 conv2D layers with 3x3 filters and \{64, 128, 256\} output channels, each followed by max pooling of stride 2, and then two fully connected layers with \{512, 10\} output units. Right: VGG-9 is the same as the VGG-11 used in \cite{simonyan-2014-arXiv-very-deep-convolutional} except with the last 2 conv2D layers removed, half the output channels in the remaining conv2D layers, and \{512, 512, 10\} output channels on the fully connected layers.
}

\figgp{resnet_firstlayer_crop.pdf}{0.69}{layer_trajs_resnet_00_crop.pdf}{0.3}{
  Left: LCA summed over all of training and across each layer of CIFAR--ResNet on SGD. Bias and batch norm layers are combined into their corresponding kernel layers. Blue represents a normal run configuration, and other colors show various experiments on the first layer.
  When the first layer uses a 10x smaller learning rate than the other layers (orange), per-layer LCA does not change much.
  While the ``first layer frozen'' runs (green) no longer hurt in the first layer (since the layer parameters are frozen from the beginning), the other layers, especially the next two, do not help as much.
  A similar effect is seen when we freeze the first layer at its LCA argmin (red); while we force the first layer to have negative LCA, the others have slightly more positive LCA, thus cancelling out any improvements.
  Middle: resulting train loss for each run configuration and standard deviations. Right: a typical cumulative trajectory of the first layer's learning, which helps in the first few hundred iterations and then increasingly hurts. The ``freeze first layer at argmin'' lets the layer help first before freezing it, but that still doesn't improve performance.
  % layer_traj plot is from may1_resnetfreeze/resnet_sgd0.1normal_0
}

\figp[h!]{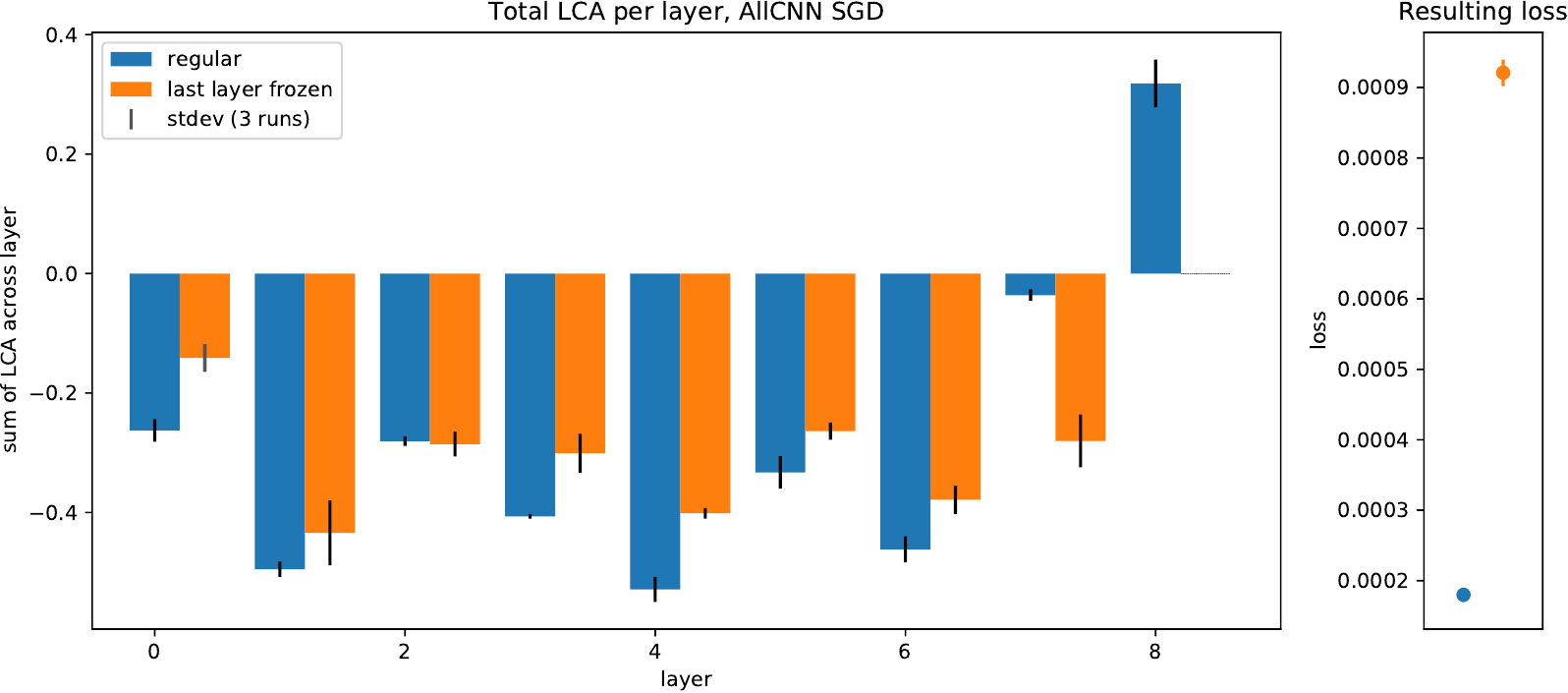}{0.8}{
  LCA for AllCNN layers. Last layer hurts in a regular run (blue), but freezing the last layer at initialization (orange) results in a worse overall loss (shown on the right).
}

\clearpage
\section{Supplementary results: Learning is synchronized across layers}
\seclabel{si:synchronized}

% There are 22.5 lines on average and their appearance statistically significant with a p-value of $8.16e^{-6}$
% There are 6.8 lines on average and their appearance are statistically significant with a p-value of $1.03e^{-9}$. 
\figsp{resnet_worm_plot_vf_highlight_crop}{.95}{Peak learning iterations by layer by class on CIFAR--ResNet. We consider the first 400 training iterations, by which point the network achieves a test accuracy of 65\%. We plot the top 20 iterations by LCA for each class and each layer, where that iteration represents a local minimum for LCA. The layers are ordered from bottom to top. Points highlighted in orange represent iterations where 25\% to 50\% of the kernel layers (6 to 10) had peak learning for that particular class, and there are 16.6 lines on average.
  Points highlighted in red represent iterations where at least 50\% of the kernel layers (11 or more) had peak learning for that particular class, and there are 5.8 lines on average.
}

\figp{fc_worm_plot_eff_grads_vf_highlight_crop}{0.85}{Peak effective gradient iterations by layer by class on MNIST--FC. 
}

\clearpage % MOVE IF NEEDED

\figp{fc_aggregate_worm_plot_vf_highlight_crop}{.92}{Peak learning iterations by layer on MNIST--FC. Eeah row represents a layer in the 3-layer FC network, orderred from bottom to top. Dots indicate top 20 moments of learning. and marked in red whenever synchronized across all layers. In this example 12 out of 20 moments are synchronized. The number of synchronized learning iterations is significantly more than chance, with a p-value of <0.001.
}

\figp{fc_aggregate_worm_plot_eff_grads_vf_highlight_crop}{0.92}{
    Peak effective gradients iterations by layer on MNIST--FC. 
}

\figp[h!]{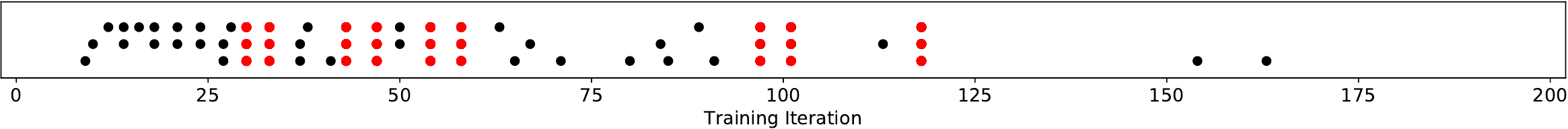}{.92}{Peak weight movement iterations by class on MNIST--FC.
}

%%%%%%%%%%%%%%%%%%%%%%%%%%%%%
% \clearpage
\section{Supplementary results: Additional observations}
\seclabel{si:learn_together}

\subsection{Trajectory of parameters and temporal correlations}
    Our method allows for each individual parameter to have its own ``loss curve'', or, cumulative LCA, as seen in \figref{si_lenetadam_params_00_crop.pdf}. Interestingly, parameters going into the same neuron tend to have similar loss curves, meaning that they learn together. This can also be seen in the animations in \figref{3frames_fc.pdf}. We prove this concept with a correlation metric and statistical test. 
    We conduct experiments for every layer in a network (focusing on kernel parameters, or weights, and ignoring bias and batch normalization parameters), calculate correlation coefficients of pairs of parameters, and apply Kolmogorov-Smirnov test to measurements for statistical significance. We find significantly stronger correlations between parameters of the same inputs or outputs than a random baseline, as depicted in \figref{input_output_correlations-crop}.

    \removed{Lots of text from before, and the figure of parameter and neuron trajectories because we already have parameter trajectories... See comments in main.tex}

    \figp{input_output_correlations-crop}{.8}{Correlation of weights within inputs and outputs, for every kernel layer of CIFAR-ResNet. \textbf{(Left)} Schematic indicating how weights belonging to the same input/output is like. \textbf{(Right)} Measured correlations for each layer. Multiple lines indicate multiple runs. For each layer, for each input/output, take all the weights going belonging to it, calculate pairwise correlation coefficients and the average of them. Then average through all nodes of that layer. Baseline for it is a constructed "fake node" with the same number of weights (or the most that exist), where no pair is from the same input or output. }

\subsection{Class specialization in neurons }

%\todo{Would a citation be needed?}
It is generally known that earlier layers in a neuron network learn more general concepts than later layers. This is akin to measuring the degree to which individual neurons specialize in learning specific classes. We can now show precisely how specialized neurons are and how this pattern evolves as we go deeper in the network.

\figp{neuron_specialization_vf.pdf}{.9}{
  \capsec{(Left)} MNIST--FC. \capsec{(Right)} CIFAR--ResNet. Fraction of neurons that concentrate on learning 1, 2, or 3 classes. For each neuron in each layer, we compute the ratio between amount helped for the top 1, 2, or 3 class(es) and the total amount helped for all positively benefited classes. We then find the fraction of neurons in each layer where the top 1, 2, or 3 class(es) contributed more than 80\% of total learning.}

We can also identify neurons that specialize in certain classes and visualize their behavior. For example, we look at two neurons in \figref{neuron_examples_vf.pdf} that concentrated on learning one and two classes, respectively. A saliency map using amount helped for each neuron gives us some insight into what that neuron is learning. A similar plot using just the weight values do not hold a clear pattern.

\figp[h!]{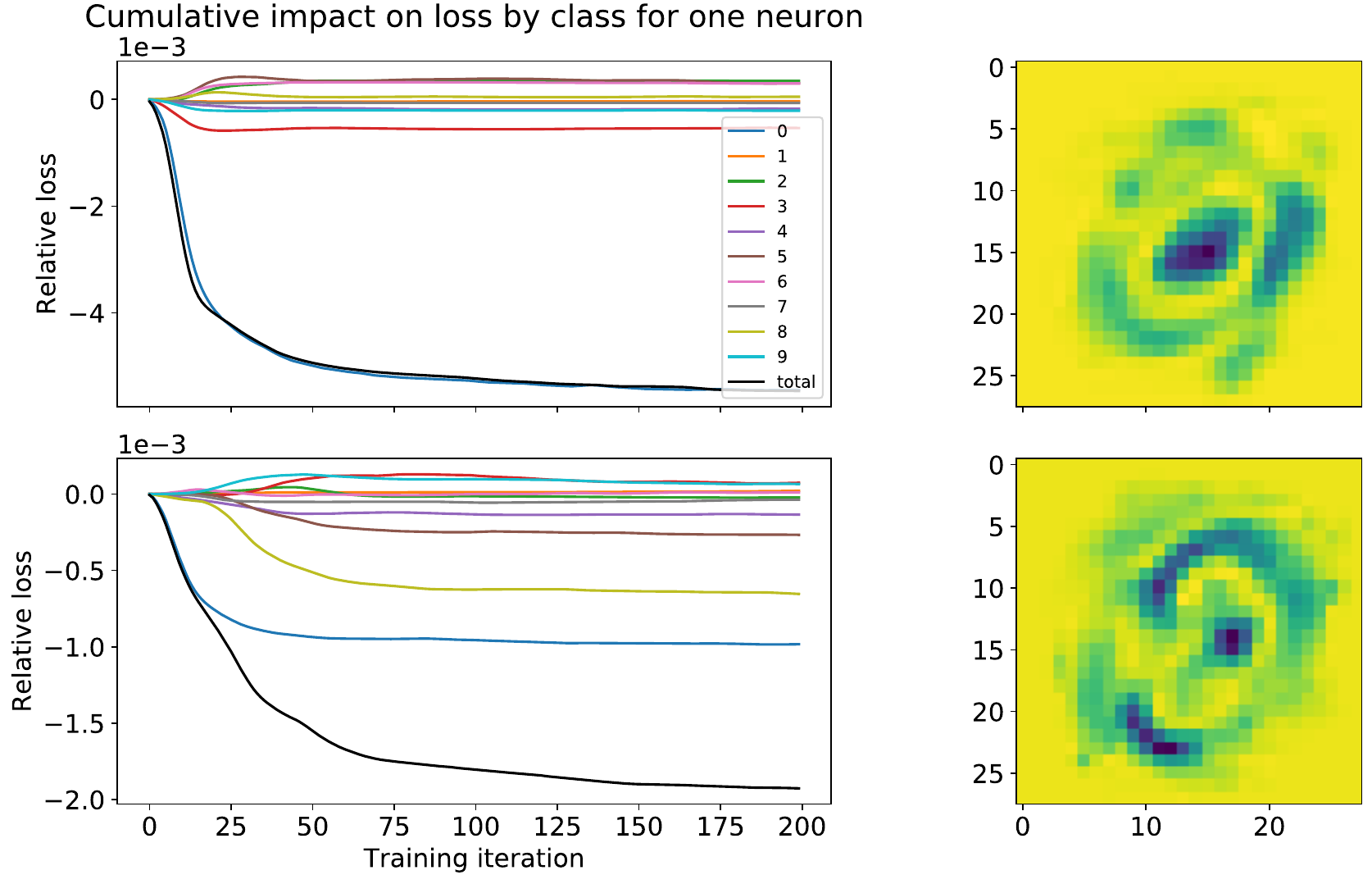}{0.6}{
  \capsec{(Top)} An example of a first layer neuron that concentrated on learning the 0 class. \capsec{(Bottom)} An example of a first layer neuron that concentrated on learning the 0 and 8 classes. \capsec{(Left)} Cumulative amount helped by class. \capsec{(Right)} Plot of the 784 parameters within this neuron reshaped as a 2D image, colored by the LCA for each parameter.}

\clearpage
\section{Additional details on model architectures and training hyperparameters}
\seclabel{si:experiment}

%
%as well as LeNet \cite{lecun-1998-IEEE-gradient-based-learning-applied}. %for an example of convolution, with two conv2D layers with max pooling followed by two fully connected layers with dropout \cite{hinton2012improving-neural-networks-by-preventing}.
%For CIFAR, we use AllCNN \cite{springenberg-2014-arXiv-striving-for-simplicity:-the-all-convolutional} with 9 convolutional layers followed by global average pooling. Batch normalization~\cite{ioffe-2015-arXiv-batch-normalization:-accelerating} and dropout~\cite{hinton2012improving-neural-networks-by-preventing} are used. We also use a deep residual network, ResNet-20 \cite{he-2015-arXiv-deep-residual-learning}.
%%as described in [\url{https://github.com/keras-team/keras/blob/master/examples/cifar10_resnet.py} and cite \url{https://arxiv.org/pdf/1512.03385.pdf}].

All layers in network are followed by ReLu nonlinearity, and weights are initialized according to the He-normal distribution \cite{he2015delving}.
\removed{See \texttt{network\_builders.py} in the attached code for exact implementation details.} 
\removed{Full implementation code is attached with submission and will be released to the public for reproducibility.}
Exact implementation details can be found in our public codebase at \url{https://github.com/uber-research/loss-change-allocation}.

\paragraph{MNIST--FC} We use a three-layer fully connected network, of sizes 100, 50, 10. For simplicity, no batch normalization or dropout was used.

\paragraph{MNIST--LeNet} First conv layer is 5x5, 20 filters, followed by a 2x2 max pool of stride 2. Second conv layer is also 5x5, 40 filters, followed by a 2x2 max pool of stride 2 and dropout of 0.25. The result is flattened and fed through two fully connected networks with output sizes 400 and 10.

\paragraph{CIFAR--AllCNN} AllCNN is the same as described in \cite{springenberg-2014-arXiv-striving-for-simplicity:-the-all-convolutional}, with 9 convolutional layers followed by global average pooling. Batch normalization~\cite{ioffe-2015-arXiv-batch-normalization:-accelerating} and dropout~\cite{hinton2012improving-neural-networks-by-preventing} are used. The table below lists the size of layers and where batch normalization and dropout are added.

\begin{tabular} { | l | l |}
\hline
3x3 conv, 96 & followed by batch normalization \\ \hline
3x3 conv, 96 & followed by batch normalization \\ \hline
3x3 conv, 96 & Stride 2, followed by 0.5 dropout \\ \hline
3x3 conv, 192 & followed by batch normalization \\ \hline
3x3 conv, 192 & followed by batch normalization \\ \hline
3x3 conv, 192 & Stride 2, followed by 0.5 dropout \\ \hline
3x3 conv, 192 & followed by batch normalization \\ \hline
1x1 conv, 192 & followed by batch normalization \\ \hline
1x1 conv, 10 & \\ \hline
Global average pooling & \\ \hline
\end{tabular}

\paragraph{CIFAR--ResNet20}
Each residual block consists of two 3x3 conv layers with the specified number of filters. Shortcuts are added directly if the number of filters is the same between blocks, otherwise the dimension change is done by a 1x1 conv with stride 2.

\begin{tabular}{ | l |}
\hline
3x3 conv, 16 \\ \hline
[Residual block of 16] x 3\\ \hline
[Residual block of 32] x 3\\ \hline
[Residual block of 64] x 3 \\ \hline
Global average pooling \\ \hline
Fully connected, 10 outputs \\ \hline
\end{tabular}

\paragraph{Hyperparameter search}
We adjusted learning rate based on validation accuracy. The range we tried is an approximate log scale [1, 2, 5] times different powers of ten from 0.0005 to 1. Early stopping iteration was selected by when validation accuracy has flattened and train is mostly complete. For momentum, batch size, and dropout rates, we used reasonable and common values. We did not tune these (unless noted in special experiments) as they worked well and the exact values were not important. We used the given training/validation/testing split in the MNIST and CIFAR datasets. 

For the LCA method, we also tried trapezoid rule, midpoint rule, and Boole's rule, and found that the Runge-Kutta method worked best given the same amount of computation.

\paragraph{Learning rates used} The following learning rates are used in default experiments (SGD uses 0.9 momentum if not otherwise stated):

\begin{tabular} { | l | c | c |}
\hline
& SGD & Adam \\ \hline
MNIST--FC, no momentum & 0.5 & N/A \\ \hline
MNIST--FC & 0.05 & 0.002 \\ \hline
MNIST--LeNet & 0.02 & 0.002 \\ \hline
CIFAR--ResNet & 0.1 & 0.005 \\ \hline
CIFAR--AllCNN & 0.1 & 0.001 \\ \hline
\end{tabular}

\section{Computational considerations}
\seclabel{comptime}

Consider the two terms of \eqnref{two}.
The second term, $\theta_{t+1}^{(i)}-\theta_{t}^{(i)}$, depends on the path taken by the optimizer through $\theta$ space, which in turn depends only on the gradients of mini-batches from the training set. This term is readily available during typical training scenarios without requiring extra computation.
In contrast, the first term, $\nabla_\theta L(\theta_{t})$, is computed over the entire training set
%or test set
and is not available in typical training. It must be computed separately, and because it requires evaluation of the entire training set, this computation is expensive (if the entire training set is $N$ times larger than your mini-batch, each iteration would take approximately $N$ times as long). Evaluating the loss and gradients over the entire training set at every training iteration may seem intractably slow, but in fact for smaller models and datasets, using modern GPUs we can compute this gradient in a reasonable amount of time, for example, 0.2 seconds per gradient calculation for a simple fully connected (FC) network on the MNIST dataset or 9 seconds for the ResNet-20 \cite{he-2015-arXiv-deep-residual-learning} model on the CIFAR-10 dataset. These times are quoted using a single GPU, but to speed calculations we distributed gradient calculations across four GPUs (we used NVIDIA GeForce GTX 1080 Ti).
Thus, although the approach is slow, it is tractable for small to medium models.

\begin{center}
\begin{tabular}{ | p{2.6cm} | p{1.8cm} | p{2.7cm} | p{2.7cm} | p{2.0cm} | } 
\hline
Model & Number of trainable params & Training time and iterations used & Time per iteration of gradient calculations on one GPU & Storage space per iteration \\ 
\hline
MNIST--FC & 84,060 &  4 min,            & 0.4 s  & 300 kb \\
          &        &  880 iterations   &        & \\
\hline
MNIST--LeNet & 808,970 & 11-12 min     & 2.4 s & 3.5 mb \\ 
          &        &  880 iterations   &        & \\
\hline
CIFAR--ResNet20 & 273,066 & 120-125 min, & 18-20 s & 1.2 mb \\
          &        &  5000 iterations    &        & \\
\hline
CIFAR--AllCNN & 1,371,658 & 270-280 min, & 36-42 s & 6 mb \\
          &        &  5000 iterations   &        & \\
\hline
\end{tabular}
\end{center}

\end{document}